\documentclass[journal,compsoc]{IEEEtran}
\usepackage[breaklinks=true,colorlinks,citecolor=black,linkcolor=black,urlcolor=black]{hyperref}
\usepackage[dvipsnames,table,xcdraw]{xcolor}

\usepackage[nocompress]{cite}
\usepackage{graphicx}
\usepackage{amsfonts}       
\usepackage{amsmath}
\usepackage{amssymb}
\usepackage{dsfont}
\usepackage{multirow}
\usepackage{bm}
\usepackage{bbm}
\usepackage[ruled]{algorithm}
\usepackage{algpseudocode}
\usepackage{overpic}
\usepackage{pifont}
\usepackage[caption=false]{subfig}

\usepackage{silence}
\newlength\savewidth

\makeatletter\renewcommand\paragraph{\@startsection{paragraph}{4}{\z@}
  {.5em \@plus1ex \@minus.2ex}{-.5em}{\normalfont\normalsize\bfseries}}\makeatother
  
\newcommand{\myPara}[1]{\vspace{.05in}\noindent\textbf{#1}\quad}

\newcommand{\figref}[1]{Fig.~\ref{#1}}%
\newcommand{\secref}[1]{Sec.~\ref{#1}}
\renewcommand{\eqref}[1]{Eq.~(\ref{#1})}

\begin{document}

\title{Zone Evaluation: Revealing Spatial Bias in Object Detection}

\author{Zhaohui Zheng,
        Yuming Chen,
        Qibin Hou,~\IEEEmembership{Member,~IEEE},
        Xiang Li,~\IEEEmembership{Member,~IEEE},
        Ping Wang,\\
        and Ming-Ming Cheng,~\IEEEmembership{Senior Member,~IEEE}
        \IEEEcompsocitemizethanks{
	\IEEEcompsocthanksitem Z. Zheng, Y. Chen, Q. Hou, X. Li, and M.M. Cheng are with VCIP, School of Computer  Science, Nankai University, Tianjin, China (Corresponding author: Qibin Hou).
         \IEEEcompsocthanksitem P. Wang is with the School of Mathematics, Tianjin University, Tianjin, China.
         \IEEEcompsocthanksitem This research was supported by NSFC (NO. 62225604, No. 62276145, No. U23B2049), the Fundamental Research Funds for the Central Universities (Nankai University, 070-63223049), and CAST through
Young Elite Scientist Sponsorship Program (No. YESS20210377). Computations were supported by the Supercomputing Center of Nankai University (NKSC). 
	}
        
        }

\markboth{IEEE TRANSACTIONS ON PATTERN ANALYSIS AND MACHINE INTELLIGENCE}%
{Shell \MakeLowercase{\textit{et al.}}: A Sample Article Using IEEEtran.cls for IEEE Journals}


\IEEEtitleabstractindextext{%
\begin{abstract}

A fundamental limitation of object detectors is that they suffer from ``spatial bias'', and in particular perform less satisfactorily when detecting objects near image borders.
For a long time, there has been a lack of effective ways to measure and identify spatial bias, and little is known about where it comes from and what degree it is.
To this end, we present a new zone evaluation protocol, extending from the traditional evaluation to a more generalized one, which measures the detection performance over zones, yielding a series of Zone Precisions (ZPs).
For the first time, we provide numerical results, showing that the object detectors perform quite unevenly across the zones.
Surprisingly, the detector's performance in the 96\% border zone of the image does not reach the AP value (Average Precision, commonly regarded as the average detection performance in the entire image zone).
To better understand spatial bias, a series of heuristic experiments are conducted.
Our investigation excludes two intuitive conjectures about spatial bias that the object scale and the absolute positions of objects barely influence the spatial bias.
We find that the key lies in the human-imperceptible divergence in data patterns between objects in different zones, thus eventually forming a visible performance gap between the zones.
With these findings, we finally discuss a future direction for object detection, namely, spatial disequilibrium problem, aiming at pursuing a balanced detection ability over the entire image zone.
By broadly evaluating 10 popular object detectors and 5 detection datasets, we shed light on the spatial bias of object detectors.
We hope this work could raise a focus on detection robustness.
The source codes, evaluation protocols, and tutorials are publicly available at \url{https://github.com/Zzh-tju/ZoneEval}.
\end{abstract}

}

\maketitle

\section{Introduction}\label{intro}
\IEEEPARstart{O}{bject} detection has shown impressive progress over the past two decades 
\cite{fasterrcnn,SSD,yolov3,DETR}.
While the optimization pipelines of object detectors are well-explored, their behaviors in a local image zone remain a mystery.
The spatial robustness of a detector is especially fundamental \cite{szabo2022mitigating} as objects may appear anywhere and all of them should be detected well.
This is particularly important for safety vision applications, e.g., fire/smoke detection \cite{huang2022fire,saponara2021real}, collision prevention in self-driving cars \cite{choi2019gaussian,badue2021self}, crowd counting and localization \cite{zhang2016single,idrees2018composition,wan2021generalized,song2021rethinking}, weapon detection in smart surveillance system \cite{bhatti2021weapon,narejo2021weapon}, and shoplifting detection \cite{kirichenko2022detection}, etc., where the border zone occupies a large proportion of the image area.
\begin{figure}[!t]
  \centering
  \includegraphics[width=0.48\textwidth]{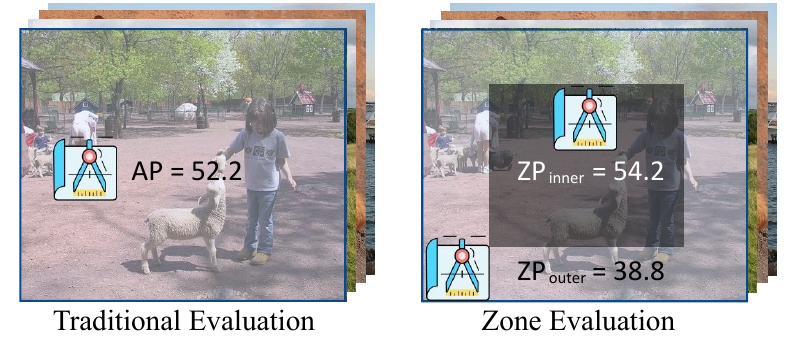}\\
  \caption{The traditional evaluation measures the detection performance over the entire image zone, but it neglects the measurement of local zone and can hardly reflect the spatial bias.
    Our zone evaluation (ZP, the average precision constrained in the zone) compensates for these issues, 
    indicating a large performance gap between zones. 
    The results are reported by GFocal \cite{gfocal} on the VOC 2007 test set \cite{voc}.
  }
  \label{fig:photographer-bias}
  \vspace{-10pt}
\end{figure}

Unfortunately, the detectors are in fact unable to perform uniformly across the spatial zones, commonly showing an evident performance drop near the image border.
This phenomenon, which we refer to as ``spatial bias'', is a natural obstacle in object detection, yet somehow being ignored for a long time by the detection community.
Setting aside the issue may result in serious safety hazards and the risk of substantial property damage.
For example, a fire detector may be good at detecting fire in the central zone while losing its ability to detect fires in the border areas of the image.
Such a fire alarm system is unreliable since the central zone of the lens only occupies a small proportion of the image area.
Some of the recent breakthroughs in spatial robustness of Convolution Neural Networks (CNNs)~\cite{azulay2019deep,zhang2019making,kayhan2020translation,chaman2021truly},
edging towards that ever-elusive translation invariance, 
have their basis for understanding how a small image transformation (e.g., color jitter, translation) does impact classification accuracy.
\begin{figure*}[!t]
  \centering
  \includegraphics[width=0.96\textwidth]{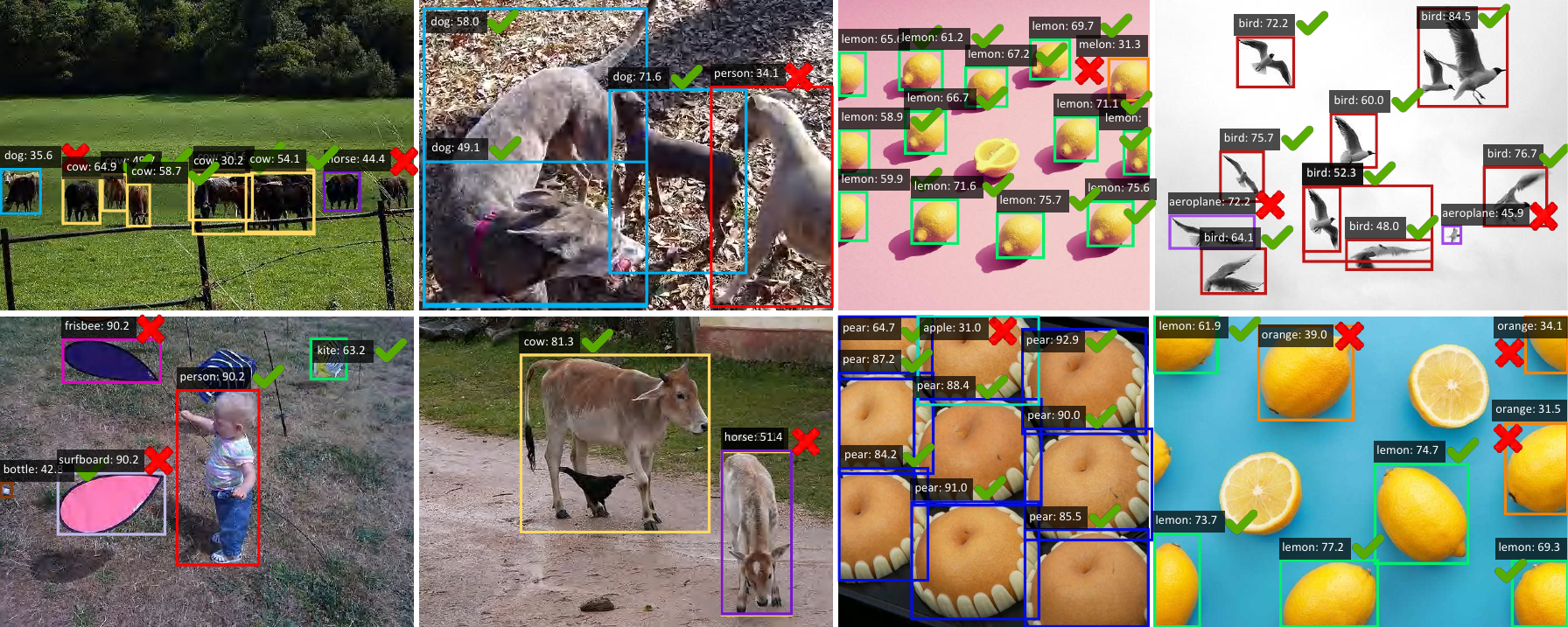}\\
  \vspace{-5pt}
  \caption{The detector is less sanguine in detecting border objects. 
    The visualizations are reported by GFocal \cite{gfocal}. Zoom in for a better view.
  }
  \vspace{-5pt}
  \label{fig:visual}
\end{figure*}
It has been found that even for the same object, the classifier can make completely different predictions as its spatial position changes \cite{kayhan2020translation}.
Beyond image classification, we in this work, delve into spatial bias in object detection, shedding light on the limitations of modern object detectors.

It has been an open issue for years that there is a lack of an effective way for measuring and identifying spatial bias.
The traditional evaluation, i.e., AP metric, measures the detection performance over the entire image zone, which provides nothing guided for the spatial robustness of detectors, and one can hardly know where and how much the performance drops.
Therefore, an evaluation protocol is particularly essential as it can provide the opportunity to better understand the spatial bias and provide tools to further build on methodology.
To this end, we present a systematic way, termed zone evaluation, to analyze whether there is spatial bias in modern object detectors, and if so, how much.

Specifically, we extend the traditional full-map evaluation to a more generalized one.
We calculate the common Average Precision (AP) \cite{coco} within a designated zone, yielding Zone Precision (ZP).
During evaluation, only the boxes whose centers lie in the zone are considered.
With these auxiliary metrics, we for the first time provide the numerical results that explicitly reveal the object detectors are in fact quite spatially biased across the zones.
As shown in \figref{fig:photographer-bias}, 
the ZP gap is 15.4 between the inner zone and the outer one.
Ostensibly, we can infer from this performance gap that the detection ability seems to be highly related to the absolute position of the object.
However, this plausible conjecture has fundamentally many inconsistencies in practice when we shift the object in an image.
We are unaware of any satisfactory explanation for the performance drop in the border zone (see \figref{fig:visual}).
Thus, we would like to shed light on the existence and the major source of spatial bias in object detectors.
Finally, we suggest a focus in future object detection research: toward spatial equilibrium.

The contributions of this work mainly involve three exploratory experiments on spatial bias, one potential research direction, and a comprehensive evaluation of modern object detectors.

\myPara{Does object scale play a key role in centralized zone performance?} (\secref{sec:4.1})
Our answer is no.
While large objects appear relatively frequently in the central zone, we observe that the zone performance remains quite uneven across different zones when we largely eliminate the effect of object scale.
It is still difficult to forge a necessary connection between spatial bias and object scale.

\myPara{Does the detector produce the centralized zone performance according to the absolute spatial position of the object?} (\secref{sec:4.2})
Our answer is no. The detection performance is barely relevant to the spatial position of objects. 
We observe that it can not statistically lead to an increase in detection quality when the objects move to the central zone.

\myPara{What will actually determine the spatial bias in object detectors?} (\secref{sec:4.3})
Our analysis reveals strong evidence that spatial bias is highly related to the discrepancy of object data patterns between the zones.
Specifically, there is a distinction in object data patterns between the central objects and the border objects.
As a consequence, an object could be better detected if it is sampled from a central-zone-style data distribution rather than a border-zone-style one.

\myPara{A new future direction: spatial disequilibrium problem.} (\secref{sec5})
The work takes a step further and presents a practical issue that urgently needs to be addressed, namely, spatial disequilibrium problem. 
In such a problem setting, the object detector incorporates spatial equilibrium as one of the important goals, which has vital significance to robust detection.
Facing this challenge, we also provide our first attempt, spatial equilibrium learning, toward spatial equilibrium object detection. (\secref{sec:5.2})

\myPara{A comprehensive evaluation.}
(\secref{sec:results})
We provide extensive evaluation and comparison of several representative object detectors.
We reveal through experiments that 1) Spatial bias is quite common in various object detectors and datasets.
2) The spatial equilibria of detectors vary significantly.
In particular, we will show that the sparse detectors perform better in the central zone, while the one-stage dense detectors perform better in the border zone.
3) The proposed spatial equilibrium learning is able to alleviate the spatial disequilibrium problem.

The remainders are organized as follows: \secref{sec:related} briefly reviews the study background.
\secref{sec3} presents zone evaluation.
\secref{sec4} studies the main origin of spatial bias.
\secref{sec5} introduces the spatial disequilibrium problem and proposes spatial equilibrium learning to alleviate this issue.
\secref{sec:results} gives extensive evaluations of zone evaluation and spatial equilibrium learning.

\section{Background}\label{sec:related}

\subsection{Relationship with data imbalance problem}
Let $\{X,G\}=\{x_i,g_i\}_{i=1}^{n}$ be a collection of sample-label pairs, 
where each sample $x_i$ has a set of ground-truth labels $g_i$.
The model training is conducted on the subset of $\{X,G\}$, 
and the network glances through the training set at each training epoch.
The data imbalance problems are usually related to the inherent properties 
of $\{X,G\}$.
In the literature of object detection, 
there are mainly two widely discussed imbalance problems.

The first one is class imbalance problem.
In this case, the sample $X$ is composed of multiple subsets 
$X_1,X_2,\cdots,X_c$ according to the class division, 
where the number of samples is imbalanced across $c$ classes, 
thereby yielding a long-tail distribution 
\cite{zhang2021deep,ouyang2016factors,li2020overcoming,wang2021adaptive}.
The class imbalance problem naturally causes imbalanced sampling 
during training, 
hindering the classification performance for those tail classes.
Re-sampling strategies \cite{Kang2020Decoupling,mahajan2018exploring} 
and cost-sensitive learning \cite{zhou2005training,cui2019class} 
are the mainstream paradigms for class re-balancing.

The second is foreground-background sampling imbalance.
This imbalance is also derived from the data itself.
A large number of anchor points are tiled on the background area, 
which are naturally sampled to be the negatives and hence dominant 
the gradient flows.
In this case, $X$ can be split into $X_{neg}$ and $X_{pos}$, 
s.t., $X=X_{neg}\bigcup X_{pos}$.
The negative samples $X_{neg}$ can be seen as the complementary set of 
the positives $X_{pos}$, 
whose ground-truth labels are ``background'' without bounding box annotations.
The solutions to this problem are similar, including re-sampling, 
e.g., OHEM \cite{OHEM}, Guided Anchoring \cite{wang2019region} 
and IoU-balanced sampling \cite{librarcnn}, as well as cost-sensitive learning, 
e.g., Focal loss \cite{lin2017focal}, GHM loss \cite{GHM}, and PISA \cite{PISA}.

In comparison, the spatial bias is also an obstacle in object detection.
Given this, we establish a novel spatial disequilibrium problem for object detection.
In this case, the sample $X$ can be divided into multiple subsets 
according to the spatial zones, just like the class division.
Generally speaking, 
spatial disequilibrium problem shares similar characteristics to class imbalance problem.
The difference is that the latter has a long-tail distribution across classes, 
while the former considers the uneven distribution of objects over spatial zones.
And we will show in \secref{sec:5.1}, that the two problems are formally equivalent to each other.

\subsection{Robustness in CNNs}
It has been widely discussed that translation invariance is not fully held by deep CNNs \cite{azulay2019deep,zhang2019making,islam2021position,kayhan2020translation,xu2021positional} since they neglect the classical sampling theorem.
A small image transformation could cause dramatic changes in predictions, thereby hindering the robustness of the classifier.
Zhang R. \cite{zhang2019making} analyzed the flaws of the max-pooling operator and proposed to inject anti-aliasing for improving the robustness of deep networks.
Lopes et al. \cite{lopes2019improving} achieved a better robustness-accuracy trade-off by presenting a patch Gaussian augmentation.

Speaking of robustness over longer spatial ranges, recent studies \cite{kayhan2020translation,alsallakhmind} reveal that CNNs can exploit the absolute positions of objects as additional information for image classification.
Islam et al. \cite{islam2021global} further extended that CNNs encode the position information based on the ordering of the channel dimensions.
In \cite{choi2021toward}, a spatially unbiased StyleGAN2 \cite{karras2020analyzing} is proposed to tackle the distorted face generation problem in the image border due to the photographer's bias in face dataset \cite{Karras_2019_CVPR}.
Gergely et al. \cite{szabo2022mitigating} empirically identified there is a drop in classification accuracy when shifting the image to make the objects closer to the image border.
Islam, M. A. et al. \cite{islam2021position} revealed that there is a boundary effect on semantic segmentation, where the vehicle segmentation quality shows a high correlation to the density of cars in a region.
Manfredi et al. \cite{manfredi2020shift} proposed a greedy approximation of AP variations by shifting the image by a few pixels to measure the translation equivariance of object detectors, but it inevitably requires several times the inference time to complete the evaluation.
In this work, we numerically quantify the generalization ability of object detectors from the perspective of the local zone for the first time, which helps us better comprehend the existence and the discrete amplitude of spatial bias.
This offers a novel analysis tool for the reliability of object detectors. 

\subsection{Evaluation from a local perspective}
Evaluation from a local perspective has been widely demonstrated to have benefits in image quality assessment (IQA) \cite{zhai2020perceptual} since the global evaluation is not aligned with the human visual system (HVS) \cite{wang2004image,sheikh2006image,ferzli2009no,liu2011visual,zhang2015application,liu2017no}.
A global value cannot reflect the spatially non-stationary model capability.
It can be dated to early research \cite{lukas1982picture} in 1982 that the authors put forward a possible improvement in quality measure if local rather than global measurements were used.
In IQA, a two-stage structure is commonly adopted.
In the first stage, image quality is evaluated locally.
The local measurement process typically produces a quality map.
To convert such quality maps into an overall quality score, a pooling algorithm is applied in the second stage of the IQA.

Wang et al. proposed Mean-SSIM \cite{wang2004image} to obtain a spatial-smooth measurement for image distortion assessment, the key of which is to compute the local SSIM for every sliding window, and then average.
3-SSIM \cite{li2009three} assigns different weights for the edges, textures, and smooth regions.
Larson et al. \cite{larson2010most} introduced a visibility-weighted local MSE to determine perceived distortion, where the image is divided into $16\times 16$ blocks with 75\% overlap between neighboring blocks.
NIQE \cite{mittal2012making} index introduces patch selection to focus on the informative image patches.
Chen et al. \cite{chen2018lip} proposed to use Landmark Distance (LMD) to focus on measuring the quality of the synthesized lip movement.
Sun et al. \cite{sun2017weighted} proposed weighted-to-spherically-uniform PSNR (WS-PSNR) to deliver different weights for different pixels.
GMSD \cite{xue2013gradient} exploits pixel-wise gradient magnitude similarity to capture image local quality.
Fan et al. \cite{fan2017structure} proposed an S-measure for salient object detection, which first divides the image into 4 square grids and assigns different weights to each local SSIM.
Bosse et al. \cite{bosse2017deep} attempted to use a CNN-based approach for learning the local image quality.
Some methods \cite{feng2008saliency,zhang2015application,gu2016saliency,zhang2017toward} incorporate saliency maps into IQA metrics since the conspicuous location can help predict the image quality perceived by human observers.
The local metrics are helpful for describing small details and structural similarity between image patches.

Although the local evaluation has been popular for decades in many assessment systems of computer vision applications, it has not been fully investigated in object detection.
Most IQA methods are mainly for pixel prediction tasks, e.g., image restoration \cite{Liang_2021_ICCV,Zamir_2022_CVPR,gu2022vqfr}, salient/camouflaged object detection \cite{Zhao_2019_ICCV,hou2019deeply,COD}, and image super-resolution \cite{7115171,Kim_2016_CVPR}.
They cannot be directly applied to instance prediction tasks, e.g., generic object detection.
Therefore, we propose zone evaluation to fill this research blank.
In addition, our work conducts a series of heuristic experiments, which provide new insights for understanding the spatial bias of modern object detectors.
We also study several shapes of zone partitions in evaluating object detectors (annular, strip, square), while previous IQA methods pay little attention to this.

\section{Zone Evaluation}\label{sec3}
In this section, we extend the traditional object detection evaluation to a more generalized zone evaluation.
Given a test image $I$ and a set of evaluation metrics $\mathcal{M}$, 
the classic evaluation methods simultaneously calculate the metrics for
all the detections and the ground-truths over the whole image.
The elements in $\mathcal{M}$ can be the COCO-style 
AP (Average Precision)~\cite{coco}, 
mAP across 10 IoU thresholds, 
or AP for the small/medium/large objects, 
which have been widely used in object detection.
These traditional metrics measure the detection performance 
over the entire image zone but consider nothing 
about the spatial robustness of object detectors.

\myPara{Zone metric.}
Let $S=\{z^1,z^2,\cdots, z^n\}$ be a zone partition s.t. $I=\bigcup\limits_S z^i$ and $z^i\cap z^j=\emptyset$, $\forall z^i,z^j\in S$, $z^i\neq z^j$.
We measure the detection performance for a specific zone $z^i$ by only considering the ground-truth objects and the detections whose centers lie in the zone $z^i$.
Then, for an arbitrary evaluation metric $m\in \mathcal{M}$, the evaluation process stays the same as the conventional ways, yielding $n$ zone metrics, each of which is denoted by $m^i$.
We call $m^S=\{m^1,m^2,\cdots, m^n\}$ the series of zone metrics for the zone partition $S$.

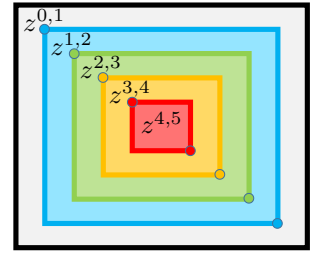
\begin{figure}[!t]
	\centering
		\begin{overpic}[width=0.45\linewidth]{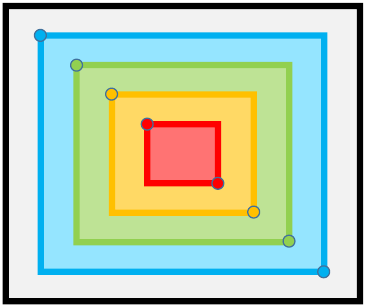}
        \put(4,73.8){$z^{0,1}$}
        \put(13,65.8){$z^{1,2}$}
        \put(22.8,58){$z^{2,3}$}
        \put(32.4,49.8){$z^{3,4}$}
        \put(43,39.6){$z^{4,5}$}
  \end{overpic}
	\caption{Definition of evaluation zone when $n=5$.}
  \label{fig:zone}
    \vspace{-5pt}
\end{figure}

\myPara{Annular zones.}
In practice, the centralized photographer's bias is ubiquitous in visual datasets \cite{torralba2011unbiased,oksuz2020imbalance,voc,coco,openimages,objects365}.
If one aims at a detector with comprehensive detection ability for all directions, the evaluation zones can be designed into a series of annular zones:
\begin{equation}\label{eq:definition-of-zone}
    z^{i,j} = R_i\setminus R_j, \quad i<j,
\end{equation}
where $R_i$ denotes a centralized region, which is given by:
\begin{equation}\label{eq:zone-def}
    R_i = \mathrm{Rectangle}((r_iW,r_iH),((1-r_i)W,(1-r_i)H)),
\end{equation}
where $\mathrm{Rectangle}(p,q)$ represents the rectangle region with the top-left coordinate $p$ and the bottom-right coordinate $q$.
$W$ and $H$ denote the width and the height of the image and $r_i=\frac{i}{2n}, i\in\{0,1,\cdots,n\}$ controls the sizes of rectangles.
An illustration of the evaluation zones can be seen in \figref{fig:zone}, where $n=5$.
We denote the average precision (AP) in the zone $z^{i,j}$ as ZP$^{i,j}$.
In this way, the traditional evaluation is a special case in our zone evaluation as one can easily get $\text{AP}=\text{ZP}^{0,n}$.

\myPara{Other zone partition.}
The traditional evaluation is flexible to select different IoU thresholds for different application scenarios.
For those requiring precise box localization, a rigorous IoU threshold can be selected, e.g., $IoU=0.75$ (AP$_{75}$).
For those less requiring box localization, AP$_{50}$ is good enough.
For example, rotated object detection methods usually report AP$_{50}$~\cite{GWD,KLD,DCL}.
Analogous to the AP metric, users can flexibly design various zone partitions based on their own applications.
If the user cares about the comprehensive detection ability for all directions, the annular zone partition would be a good choice. 
If the user cares about some regions of interest, the evaluated zones can be customized.
In \secref{sec:results}, we will show the evaluation results on two more special zone partitions.
One is the strip zones, i.e., 5 zones along x-axis and 5 zones along y-axis (See \secref{sec:6.3} ``Observations 3''), and the other is square zones of $11\times 11$ blocks (See \secref{sec:6.4} ``Correlation with object distribution''.
Importantly, as the zone partition holds consistently, the comparison among the detectors stays fair.
This property helps us observe the performance of different detectors in the regions of interest so that we can choose detectors based on practical application needs.
In \secref{sec:results}, we will show that the sparse detectors perform better in the central zone, while the one-stage dense detectors perform better in the border zone.

\begin{figure*}[!t]
	\centering
		\begin{overpic}[width=0.92\textwidth]{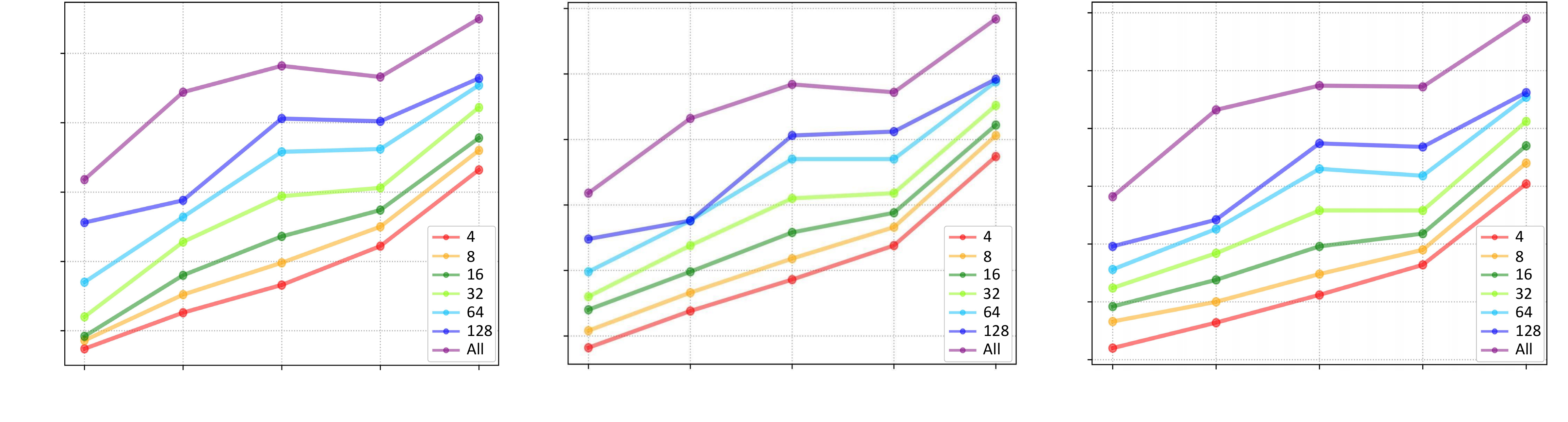}
		\scriptsize
		\put(2,5.3){20}
		\put(2,9.6){25}	
		\put(2,14.1){30}
		\put(2,18.5){35}
		\put(2,23){40}
		\put(34,5.0){20}
		\put(34,9.2){25}	
		\put(34,13.4){30}
		\put(34,17.5){35}
		\put(34,21.7){40}
		\put(34,25.9){45}
		\put(67.5,3.5){20}
		\put(67.5,7.1){25}	
		\put(67.5,10.8){30}
		\put(67.5,14.6){35}
		\put(67.5,18.2){40}
		\put(67.5,22.0){45}
            \put(67.5,25.6){50}

		\put(4,1.5){ZP$^{0,1}$}
            \put(10.2,1.5){ZP$^{1,2}$}
            \put(16.4,1.5){ZP$^{2,3}$}
            \put(22.9,1.5){ZP$^{3,4}$}
            \put(29.1,1.5){ZP$^{4,5}$}

		\put(36,1.5){ZP$^{0,1}$}
            \put(42.5,1.5){ZP$^{1,2}$}
            \put(49.1,1.5){ZP$^{2,3}$}
            \put(55.6,1.5){ZP$^{3,4}$}
            \put(62.1,1.5){ZP$^{4,5}$}

		\put(69.4,1.5){ZP$^{0,1}$}
            \put(76.1,1.5){ZP$^{1,2}$}
            \put(82.6,1.5){ZP$^{2,3}$}
            \put(89.2,1.5){ZP$^{3,4}$}
            \put(95.8,1.5){ZP$^{4,5}$}
            
		\put(14.2,-1){(a) GFocal}
		\put(43,-1){(b) Cascade R-CNN}
\put(77,-1){(c) Deformable DETR}
		\end{overpic}
	\caption{Mean ZP with various object scale ranges. One can see that for each range of the object scale $r$, the spatial bias is significant on the three object detectors.}
  \label{fig:object-scale-range}
\end{figure*}
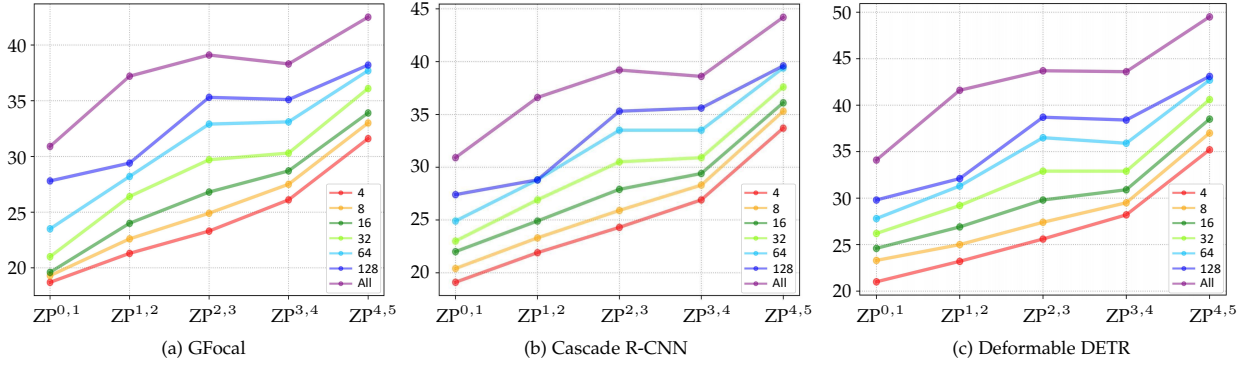
\myPara{Measuring the discrete amplitude for zone metrics.} As the detection performance varies across the zones, we further introduce an additional metric to gauge the discrete amplitude among the zone metrics.
Given all the zone metrics $m^S$ for a specific zone partition $S$, we calculate the variance $\sigma(m^S)$ of the zone metrics.
Ideally, if $\sigma(m^S)=0$, the generalization ability of the object detector reaches perfect spatial equilibrium \textbf{under the current zone partition}.
In this situation, an object can be well detected without being influenced by its data pattern.
It is also worth mentioning that \textbf{spatial bias is an external manifestation of object detectors}, and the ZP variance can only reflect the spatial equilibrium for the given zone partition.
In other words, there are three concepts as follows.

1) Let $S$ be a zone partition, the spatial equilibrium of the detector is good for $S$ if $\sigma(m^S)$ is sufficiently small.

2) Let $S_1,S_2$ be two different zone partitions with $n$ zones. If $\sigma(m^{S_1})<\sigma(m^{S_2})$, then the detector is considered to be more spatially equilibrated in the way of zone partition $S_1$.

3) The detector has no spatial bias, indicating $\forall S$ be a zone partition, $\sigma(m^S)$ is sufficiently small.

\section{Delving Deep into Spatial Bias}\label{sec4}

In this section, we conduct exploratory experiments to clarify the spatial bias about its existence and the major source.
We adopt three representative object detectors.
The first one is the popular one-stage dense object detector GFocal \cite{gfocal}.
The second is the classic multi-stage dense-to-sparse object detector Cascade R-CNN \cite{cascadercnn}.
The third is the sparse object detector Deformable DETR \cite{deformabledetr}.

\subsection{Study on object scale}\label{sec:4.1}

From the definition of evaluation zones (\eqref{eq:definition-of-zone}), one may ask whether the object scale plays a key role in centralized zone performance.
In this experiment, the annular zone partition is adopted as shown in \figref{fig:zone}.
An object belongs to $z^{i,j}$ if its central point coordinates are in $z^{i,j}$.
To eliminate the effect of object scale, we restrict the zone evaluation process in the objects with a similar scale.
For each range of object scale $r$, we select all the ground-truth boxes whose areas are in the range of $[0,r^2]$, $[r^2,(2r)^2]$, ..., $[(kr)^2,\infty]$, respectively,
where the maximum endpoint of the scale is set to $kr=256$, and $r\in\{4,8,16,32,64,128,\infty\}$.
Then, we calculate the mean value of ZP over all the scales, which is shown in \figref{fig:object-scale-range}.
We observe that the spatial bias is significant no matter how small the range of the object scale is chosen.
The ZP$^{4,5}$ score continues to be the best and in contrast, the ZP$^{0,1}$ score is the worst.
It shows a steeper drop-off in performance as the evaluation zones get closer to the image border.
One can see that the performance gap between the inner zone and the outer zone is consistently large, with more than 10 ZP gap.
This indicates that the centralized spatial bias is probably not derived from the object scale factor.
For simplicity, we adopt all scales for zone evaluation in the following experiments.

\subsection{Study on the absolute spatial position of objects}\label{sec:4.2}

\begin{figure}[!t]
	\centering
		\begin{overpic}[width=0.48\textwidth]{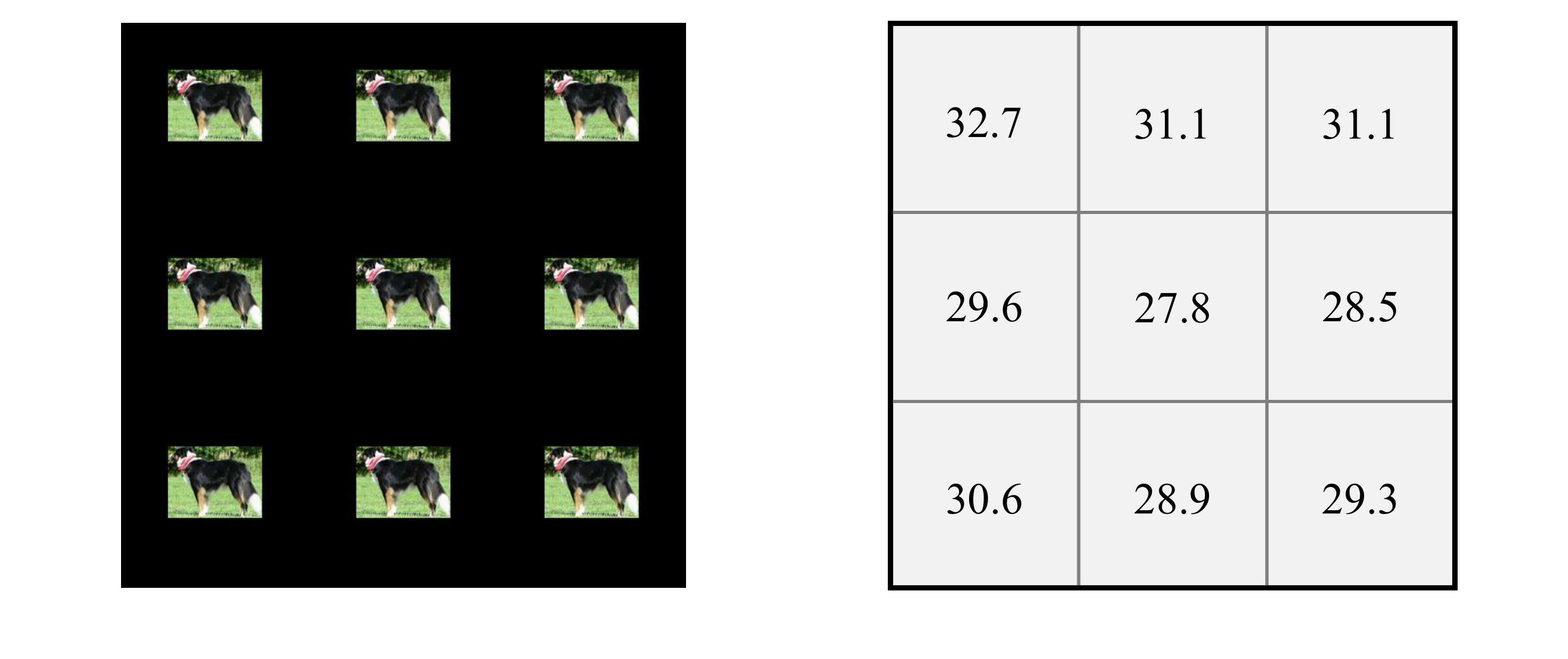}
		\scriptsize
		\put(10,-1){(a) Sudoku-style dataset.}
		\put(63,-1){(b) ZPs of 9 zones.}
		\put(64,29){$z^{11}$}
		\put(76.2,29){$z^{12}$}
		\put(88.1,29){$z^{13}$}
		\put(64,17){$z^{21}$}
		\put(76.2,17){$z^{22}$}
		\put(88.1,17){$z^{23}$}
		\put(64,5.1){$z^{31}$}
		\put(76.2,5.1){$z^{32}$}
		\put(88.1,5.1){$z^{33}$}
		\end{overpic}
	\caption{(a) The Sudoku-style dataset is constructed by regularly placing all the objects of the test set on a $600\times600$ black image. (b) Zone evaluation on GFocal ($3\times3$ grids).}
  \label{fig:spatial-position-of-objects}
\end{figure}

As the zone performance exhibits a clear centralization trend, one straightforward conjecture is that the spatial bias is related to the absolute spatial positions of objects.
The object may be better detected if it is shifted to the central zone, and reversely, be worse if it is in the border zone.
To analyze whether the spatial positions of objects play a key role in their detection quality, we construct the Sudoku-style dataset for object detection of the regularly placed objects on a black $600\times600$ image.
The experiment is based on the following 3 steps:
(1) We first crop the objects from PASCAL VOC 2007 test set \cite{voc}, 14,976 objects in total.
(2) All the objects are scaled to a fixed size and placed in a $3\times3$ grid manner. See \figref{fig:spatial-position-of-objects}(a).
(3) To measure the detection quality of each grid, the evaluation zones are defined as the same $3\times3$ zones, denoted by $z^{ij}$, $i,j\in{1,2,3}$.
It can be seen from \figref{fig:spatial-position-of-objects}(b) that the detector does not perform the best and even the worst in the central zone $z^{22}$.
This phenomenon is somewhat incompatible with previous observation in \cite{szabo2022mitigating}, which analyzes the effect of shifting 100 images and concludes that the detector may have a performance drop when the objects close to the image border.
However, when we increase the number of samples up to more than 14K, we observe that shifting objects to the central zone cannot statistically eventuate an increase in detection quality.
Therefore, it is less discernible for the relevance between the centralization tendency of detection performance and the absolute position of objects.

\textbf{Discussion:} The conclusion of \figref{fig:object-scale-range} is that the central objects can be better detected than the border objects and it is irrelevant to object scale, while the conclusion of \figref{fig:spatial-position-of-objects} is that the absolute position of the object (by object shifting) is irrelevant to forming the centralized zone performance.
We ask: What will actually determine the spatial bias in object detectors?

\subsection{Object data patterns between the zones}\label{sec:4.3}

This subsection aims to investigate the origin of the centralized spatial bias in object detectors.
Our inspiration behind it is that the centralized spatial bias may come from the differences of the object data patterns between the zones.
An object can be better detected if sampled from a central-zone-style data distribution, and worse if it is sampled from a border-zone-style one.
The representation of the object data patterns is very general \cite{bau2017network,cheng2020explaining,zhang2018examining}, and only a few adaptive modifications need to be made to the detection pipeline, requiring only 1) the input image $I$, 2) all the ground-truth boxes $G$, 3) a feature extractor $f: I\rightarrow \mathbb{R}_{M\times H\times W}$ from a pre-trained object detector.
During inference, we crop the object features from $f(I)$ by using the ground-truth boxes, and then average the feature values along the spatial dimensions.
Each object is represented by a $M$-dimensional feature vector $g$, which encodes the object data patterns in high-dimensional space.
The number of zones is set to 2 in this experiment.
We denote the central zone as $z^{in}$, which is a rectangle region with top-left coordinate $p=(0.25W, 0.25H)$ and bottom-right coordinate $q=(0.75W,0.75H)$, where $W$ and $H$ are the width and the height of the input image.
The rest part of the region is set as the border zone, denoted by $z^{out}$.
The differences of the object data patterns between the zones are represented as:
\begin{equation}
    \mathcal{E}((G_1,u),(G_2,v))= \frac{1}{KCM}\sum\limits_{k=1}^{K}\sum\limits_{c=1}^C\sum\limits_{m=1}^M\mathbbm{1}_k||\bar{g}_{m,c}^{u,G_1}-\bar{g}_{m,c}^{v,G_2}||,
\end{equation}
where $\bar{g}$ denotes the feature representation center, $u, v\in\{z^{in},z^{out}\}$ represent the objects sampled from the central zone or the border zone, as well as $G_1,G_2\in\{G_{train}, G_{test}\}$ represent the objects sampled from the train set or the test set.
The errors are calculated for each class and then averaged.
$C$ is the total number of classes.
An indicator function $\mathbbm{1}_k$ is also introduced to eliminate the effect of object scale, which is 1 when the object scale is in one of the range $R=\{[((k-1)r)^2,(kr)^2]\}\bigcup\{[((K-1)r)^2,\infty]\}$, $k\in\{0,1,\cdots,K-1\}$, and 0 otherwise.
In a nutshell, $\mathcal{E}$ measures the distance between the feature representation centers of the four sets, i.e., the central/border objects from the train/test set.

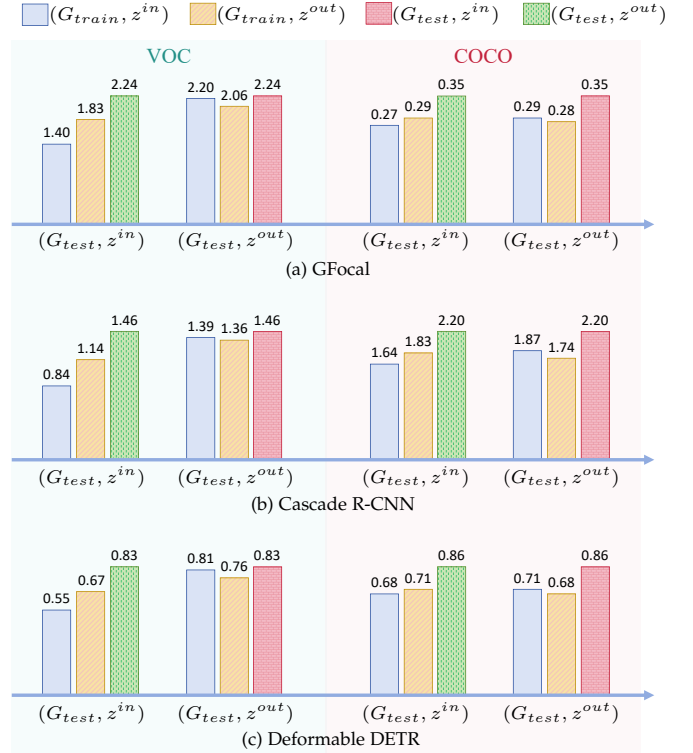
\begin{figure}[!t]
\scriptsize
	\centering
		\begin{overpic}[width=0.48\textwidth]{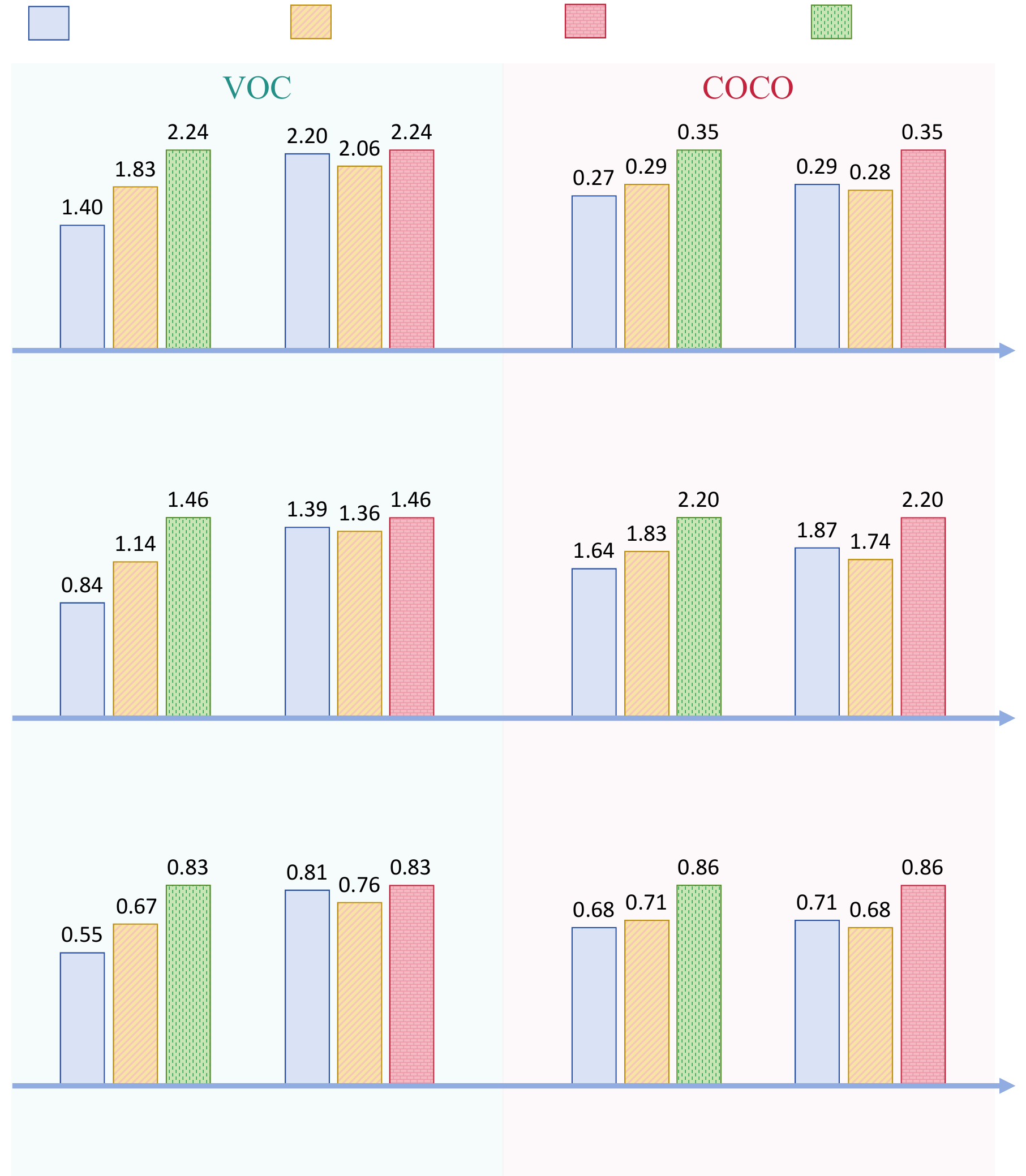}
  \put(37.4,63.4){(a) GFocal}
  \put(32.8,32.2){(b) Cascade R-CNN}
  \put(31.6,1){(c) Deformable DETR}
  
  \put(4.5,4.5){$(G_{test},z^{in})$}
  \put(23,4.5){$(G_{test},z^{out})$}
  \put(47.5,4.5){$(G_{test},z^{in})$}
  \put(66.2,4.5){$(G_{test},z^{out})$}

  \put(4.5,35.9){$(G_{test},z^{in})$}
  \put(23,35.9){$(G_{test},z^{out})$}
  \put(47.5,35.9){$(G_{test},z^{in})$}
  \put(66.2,35.9){$(G_{test},z^{out})$}

  \put(4.5,67.3){$(G_{test},z^{in})$}
  \put(23,67.3){$(G_{test},z^{out})$}
  \put(47.5,67.3){$(G_{test},z^{in})$}
  \put(66.2,67.3){$(G_{test},z^{out})$}

  \put(6,97.3){$(G_{train},z^{in})$}
  \put(28.2,97.3){$(G_{train},z^{out})$}
  \put(51.5,97.3){$(G_{test},z^{in})$}
  \put(72.5,97.3){$(G_{test},z^{out})$}
		\end{overpic}
	\caption{Average error $\mathcal{E}((G_1,u),(G_2,v))$ between the feature representation centers. For instance, the blue bar denotes $\mathcal{E}((G_{test},z^{in}),(G_{train},z^{in}))$. The objects sampled from the same zone have significantly lower differences than those sampled from different zones.}
  \label{fig:zone-error}
\end{figure}

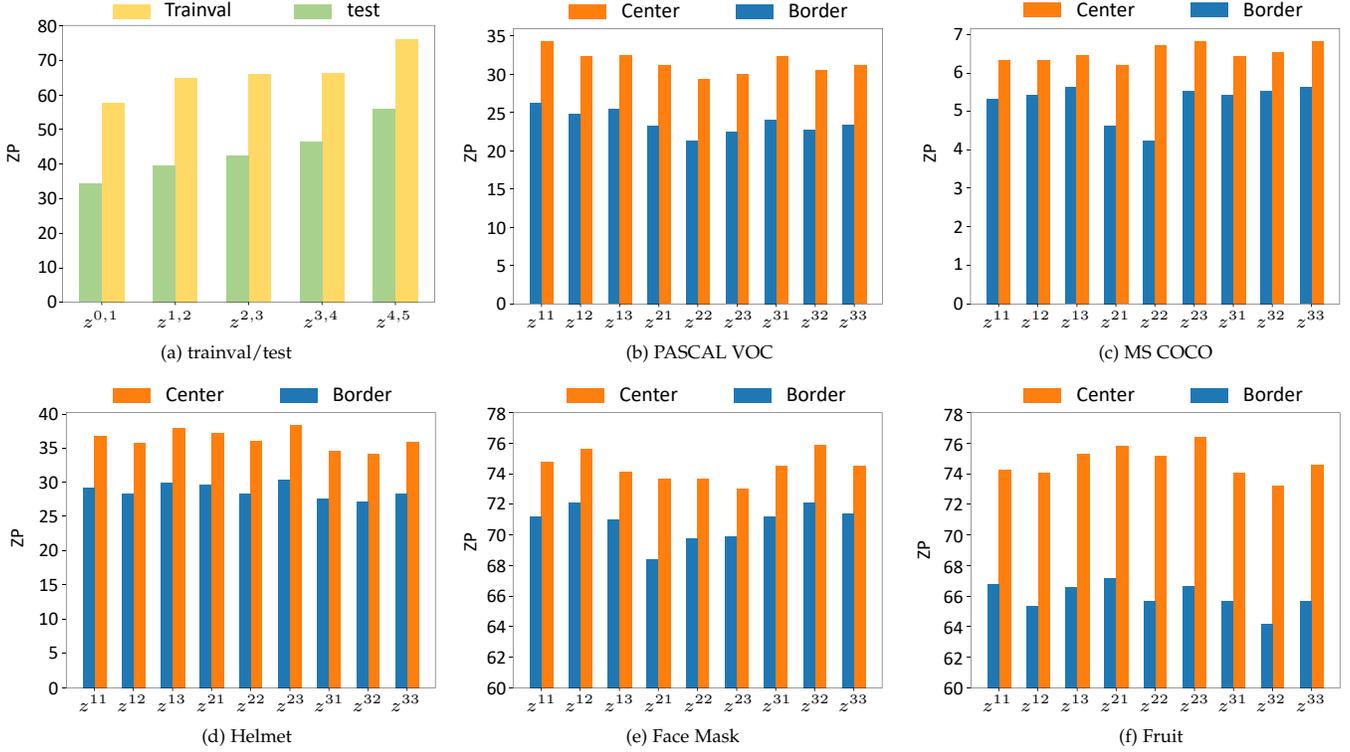
\begin{figure*}[!t]
	\centering
	\setlength{\tabcolsep}{1pt}
	\setlength{\abovecaptionskip}{3pt}
		\begin{overpic}[width=1\textwidth]{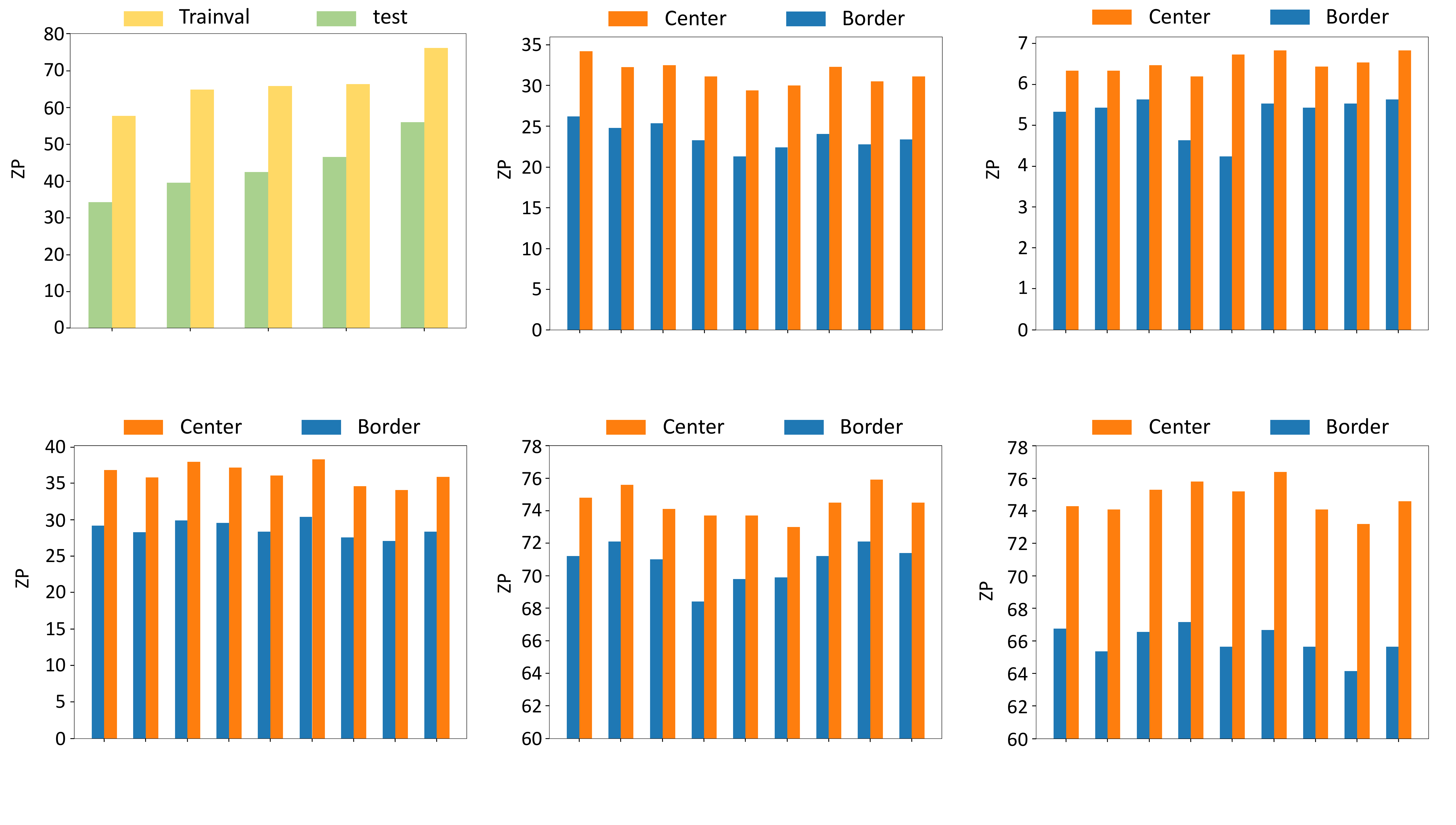}
		\scriptsize
        \put(6.3,32.3){$z^{0,1}$}
        \put(11.7,32.3){$z^{1,2}$}
        \put(17,32.3){$z^{2,3}$}
        \put(22.4,32.3){$z^{3,4}$}
        \put(27.7,32.3){$z^{4,5}$}

        \put(6,4.2){$z^{11}$}
        \put(8.8,4.2){$z^{12}$}
        \put(11.7,4.2){$z^{13}$}
        \put(14.6,4.2){$z^{21}$}
        \put(17.5,4.2){$z^{22}$}
        \put(20.4,4.2){$z^{23}$}
        \put(23.2,4.2){$z^{31}$}
        \put(26.0,4.2){$z^{32}$}
        \put(28.8,4.2){$z^{33}$}

        \put(38.7,4.2){$z^{11}$}
        \put(41.5,4.2){$z^{12}$}
        \put(44.4,4.2){$z^{13}$}
        \put(47.3,4.2){$z^{21}$}
        \put(50.2,4.2){$z^{22}$}
        \put(53.1,4.2){$z^{23}$}
        \put(55.9,4.2){$z^{31}$}
        \put(58.7,4.2){$z^{32}$}
        \put(61.5,4.2){$z^{33}$}

        \put(72,4.2){$z^{11}$}
        \put(74.8,4.2){$z^{12}$}
        \put(77.7,4.2){$z^{13}$}
        \put(80.6,4.2){$z^{21}$}
        \put(83.5,4.2){$z^{22}$}
        \put(86.4,4.2){$z^{23}$}
        \put(89.2,4.2){$z^{31}$}
        \put(92.0,4.2){$z^{32}$}
        \put(94.9,4.2){$z^{33}$}

        \put(38.7,32.3){$z^{11}$}
        \put(41.5,32.3){$z^{12}$}
        \put(44.4,32.3){$z^{13}$}
        \put(47.3,32.3){$z^{21}$}
        \put(50.2,32.3){$z^{22}$}
        \put(53.1,32.3){$z^{23}$}
        \put(55.9,32.3){$z^{31}$}
        \put(58.7,32.3){$z^{32}$}
        \put(61.5,32.3){$z^{33}$}

        \put(72,32.3){$z^{11}$}
        \put(74.8,32.3){$z^{12}$}
        \put(77.7,32.3){$z^{13}$}
        \put(80.6,32.3){$z^{21}$}
        \put(83.5,32.3){$z^{22}$}
        \put(86.4,32.3){$z^{23}$}
        \put(89.2,32.3){$z^{31}$}
        \put(92.0,32.3){$z^{32}$}
        \put(94.9,32.3){$z^{33}$}
        
        \put(12,30){(a) trainval/test}
        \put(46,30){(b) PASCAL VOC}
        \put(80.5,30){(c) MS COCO}
        \put(15,2.0){(d) Helmet}
		\put(46,2.0){(e) Face Mask}
		\put(82,2.0){(f) Fruit}
		\end{overpic} 

	\caption{(a) The 5-zone evaluation on VOC 2007 trainval set and test set. (b-f) The zone evaluation on Sudoku-style dataset with the center objects and the border objects separately. It shows that the central objects can always be better detected no matter where we place them.
		}
	\label{fig:trainval-test-center-border-ZP}
\end{figure*}

The results are reported in~\figref{fig:zone-error}. 
For VOC, the train set is VOC 2007 trainval, and the test set is VOC 2007 test.
For COCO, the train set is COCO train2017, and the test set is COCO val2017.
One can see that the objects sampled from the same zone have significantly lower differences than those sampled from different zones.
Specifically, the central objects of the test set are more similar to the central objects of the train set, and the border objects of the test set are more similar to the border objects of the train set.
This indicates that the object data patterns are actually distinct across the zones, and the network is able to capture such deviation.
As illustrated in \figref{fig:trainval-test-center-border-ZP}(a), the zone performance is centralized on VOC 2007 trainval set, thus it naturally inherits the same trend on the test set.
More intriguingly, we further visualize the detection performance on the Sudoku-style dataset in \figref{fig:trainval-test-center-border-ZP}(b-f), where the central objects and the border objects are separated. 
It can be seen that the detector can consistently perform better in detecting the central objects no matter where they are.
We notice that this phenomenon holds for all the 5 datasets, including PASCAL VOC, MS COCO, and 3 other application datasets (face mask, fruit, helmet).

The above results confirm the intuition that the objects would be better detected if they are sampled from the central-zone-style data distribution, and be worse if they are sampled from the border-zone-style one.
This shows that when we humans take photos, there is always a divergence in the object data patterns between zones, though it is imperceptible.
When the lens are focused on the regions of interest where objects are most likely to appear, it inevitably leads to a decrease in the sampling frequency of objects in the border zone, resulting in sub-optimal performance.

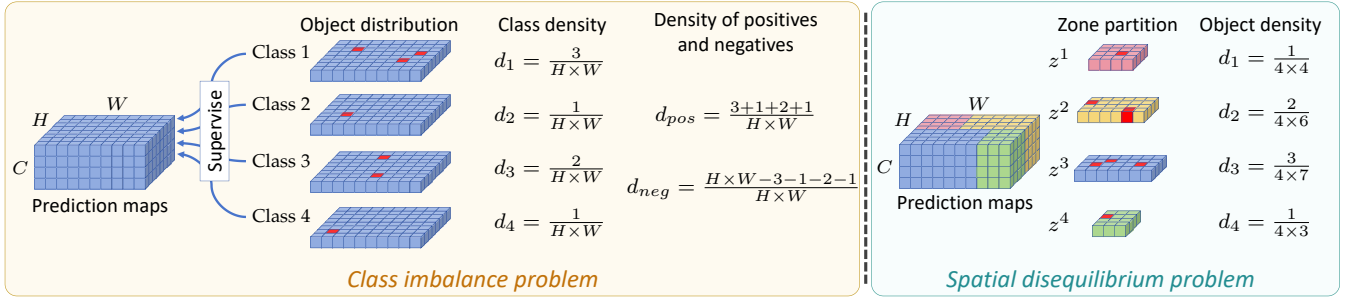
\begin{figure*}[!t]
	\centering
		\begin{overpic}[width=0.98\textwidth]{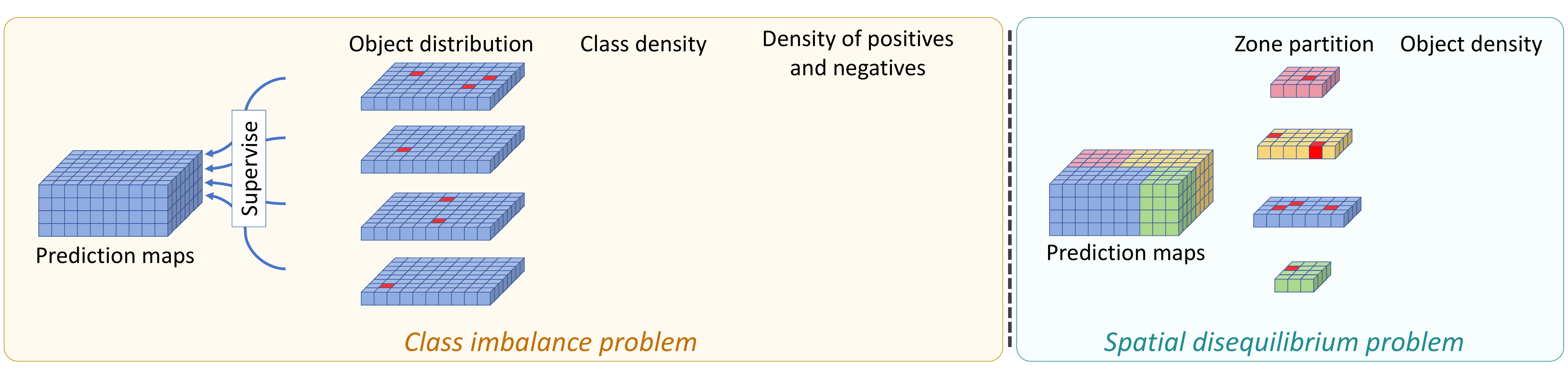}
  \scriptsize
            \put(7.8,15){$W$}
            \put(2.2,13.8){$H$}
            \put(0.8,10.2){$C$}
            \put(72.,15){$W$}
            \put(66.5,13.8){$H$}
            \put(65.2,10.3){$C$}

            \put(18.7,18.9){Class 1}
            \put(18.7,15){Class 2}
            \put(18.7,10.7){Class 3}
            \put(18.7,6.6){Class 4}

            \footnotesize
            \put(36.7,18.2){$d_1=\frac{3}{H\times W}$}
            \put(36.7,14.2){$d_2=\frac{1}{H\times W}$}
            \put(36.7,10.2){$d_3=\frac{2}{H\times W}$}
            \put(36.7,6.2){$d_4=\frac{1}{H\times W}$}
            \put(48.7,14.2){$d_{pos}=\frac{3+1+2+1}{H\times W}$}
            \put(46.6,8.8){$d_{neg}=\frac{H\times W-3-1-2-1}{H\times W}$}

            \put(78.,18.2){$z^1$}
            \put(78.,14.3){$z^2$}
            \put(78.,10.1){$z^3$}
            \put(78.,6.){$z^4$}

            \put(90.7,18.4){$d_1=\frac{1}{4\times 4}$}
            \put(90.7,14.4){$d_2=\frac{2}{4\times 6}$}
            \put(90.7,10.4){$d_3=\frac{3}{4\times 7}$}
            \put(90.7,6.2){$d_4=\frac{1}{4\times 3}$}
		\end{overpic} 
	\caption{Illustration of the relationship between class imbalance problem and spatial disequilibrium problem. There are 7 objects of 4 classes, denoted by red cubes. The evaluated zone is set to 4 zones, colored pink, yellow, blue, and green. We simplify the discussion that each object contains only 1 positive sample, and similarly for the case of multiple positives. In the left diagram, the class density is the ratio of the number of objects and the size of prediction maps for each class, whereas, in the right diagram, the object density is the ratio of the number of objects and the size of zones for each zone. The spatial disequilibrium problem is formally equivalent to the class imbalance problem.
		}
	\label{fig:two-problem}
\end{figure*}

\section{Spatial disequilibrium problem}\label{sec5}

Thus far, we have shown the existence and the major source of spatial bias.
The sub-optimal performance in the border zone impedes the robustness of detection applications.
In this section, we introduce the new spatial disequilibrium problem for object detection.

\subsection{Problem definition}\label{sec:5.1}

Denoting $S$ be a zone partition, $m^S$ be a series of zone metrics, and $\sigma: \mathbbm{R}^n\rightarrow\mathbbm{R}$ be a variance calculation, the spatial disequilibrium problem is defined to minimize the variance of the zone metrics:
\begin{equation}
\underset{\Theta}{\min}\, \sigma(m^S|\Theta),
\end{equation}
where $\Theta$ is the set of network parameters of the detector.
The general goal of facilitating spatial equilibrium is primarily decided by which zone partition to use, which is application-oriented.
Thus, for different application scenarios, the zone partition can be customized.

\myPara{Disucssion:} \textit{The spatial disequilibrium problem is formally equivalent to the class imbalance problem.}
We denote the objects in the zone $z^i$ as $X^i_{obj}$.
The density of objects for a given zone $z^i$ can be represented as $d_i=|X^i_{obj}|/|z^i|$.
Intuitively, higher density indicates more positive samples, thereby generating a larger gradient flow on the zone.
It is analogous to the class imbalance problem which has a long-tail distribution among the classes, as shown in \figref{fig:two-problem}.
Specifically, the classification branch predicts the class scores which is a $C\times H\times W$ tensor.
The density of the $c$-th class is represented as the ratio $d_c=|X_c|/(H\times W)$, $X_c$ is the object of the $c$-th class.
Since $H\times W$ is a constant, it is equivalent that all the samples of a class are placed in a zone with an identical area.
Then, each class has a single $H\times W$ zone for model learning and is disjoint among classes.
In \figref{fig:imbalanced-problem}, it can be seen that both problems are subject to a long-tail distribution.
Due to that fact, the zone performance may also be correlated to the supervision signal strength in the zone.
A simple approach is to enlarge or reduce the supervision signal strength for the zones so that the network reaches a new convergence state.
Here we plug a new parameter $\beta$ into label assignment algorithm \cite{ATSS}.
An anchor is assigned to be a positive sample if its IoU is larger than $\alpha_{pos} + \beta * \mathbbm{1}_z$, where $\alpha_{pos}$ is the positive IoU threshold, $\mathbbm{1}_z$ is 1 if the central point of this anchor lies in the zone $z$ and 0 otherwise.
Thus, the number of the positive samples $|X^i_{pos}|$ decreases when $\beta$ increases.

\begin{figure}[!t]
	\centering
		\begin{overpic}[width=0.48\textwidth]{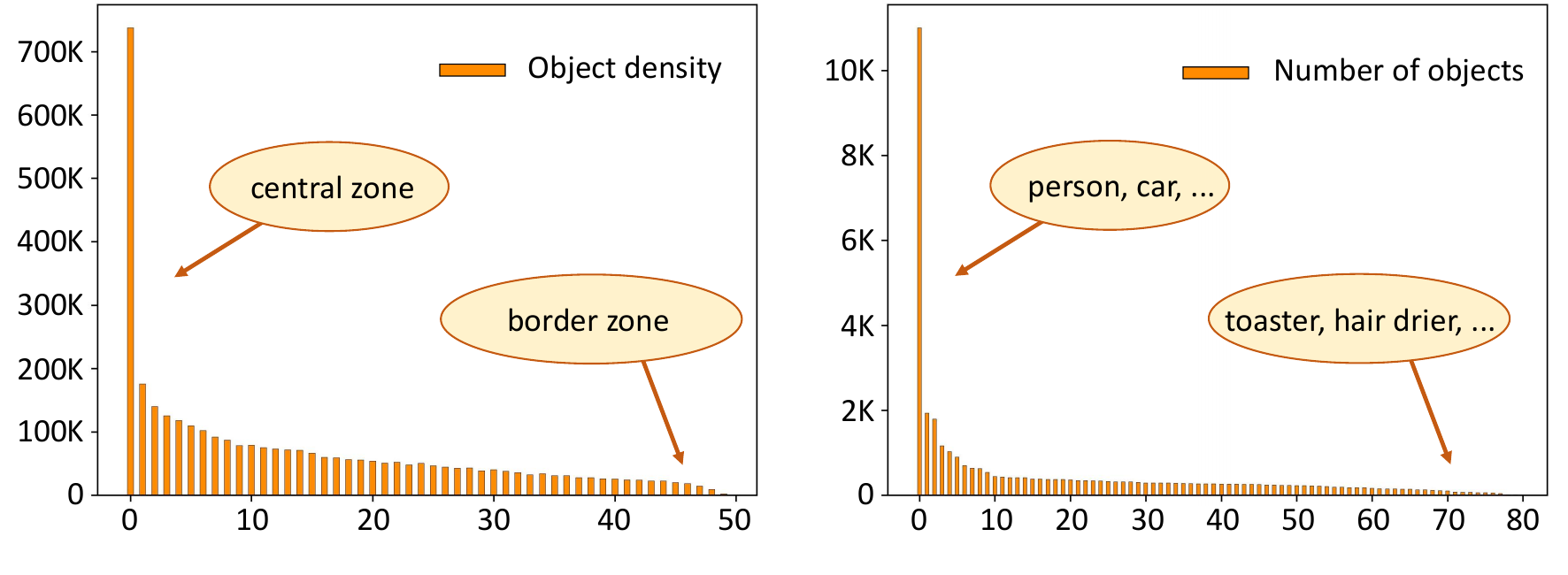}
		\put(27,-1){\footnotesize{$i$}}
		\put(75,-1){\footnotesize{class}}
		\end{overpic} 
	\caption{\textbf{Left}: The object density against 50 zones on COCO val2017. The zones are centrally defined as $z^{i,i+1}$, $i=0,1,\cdots,49$. \textbf{Right}: The long-tail distribution of classes on COCO val2017.
	The spatial disequilibrium problem shares a similar characteristic to the class imbalance problem.
		}
	\label{fig:imbalanced-problem}
 \vspace{-5pt}
\end{figure}
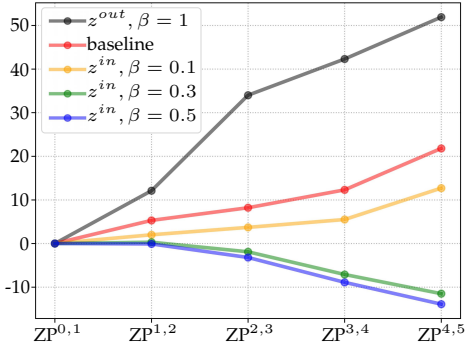
\begin{figure}[!t]
\vspace{5pt}
	\centering
        \begin{overpic}[width=0.34\textwidth]{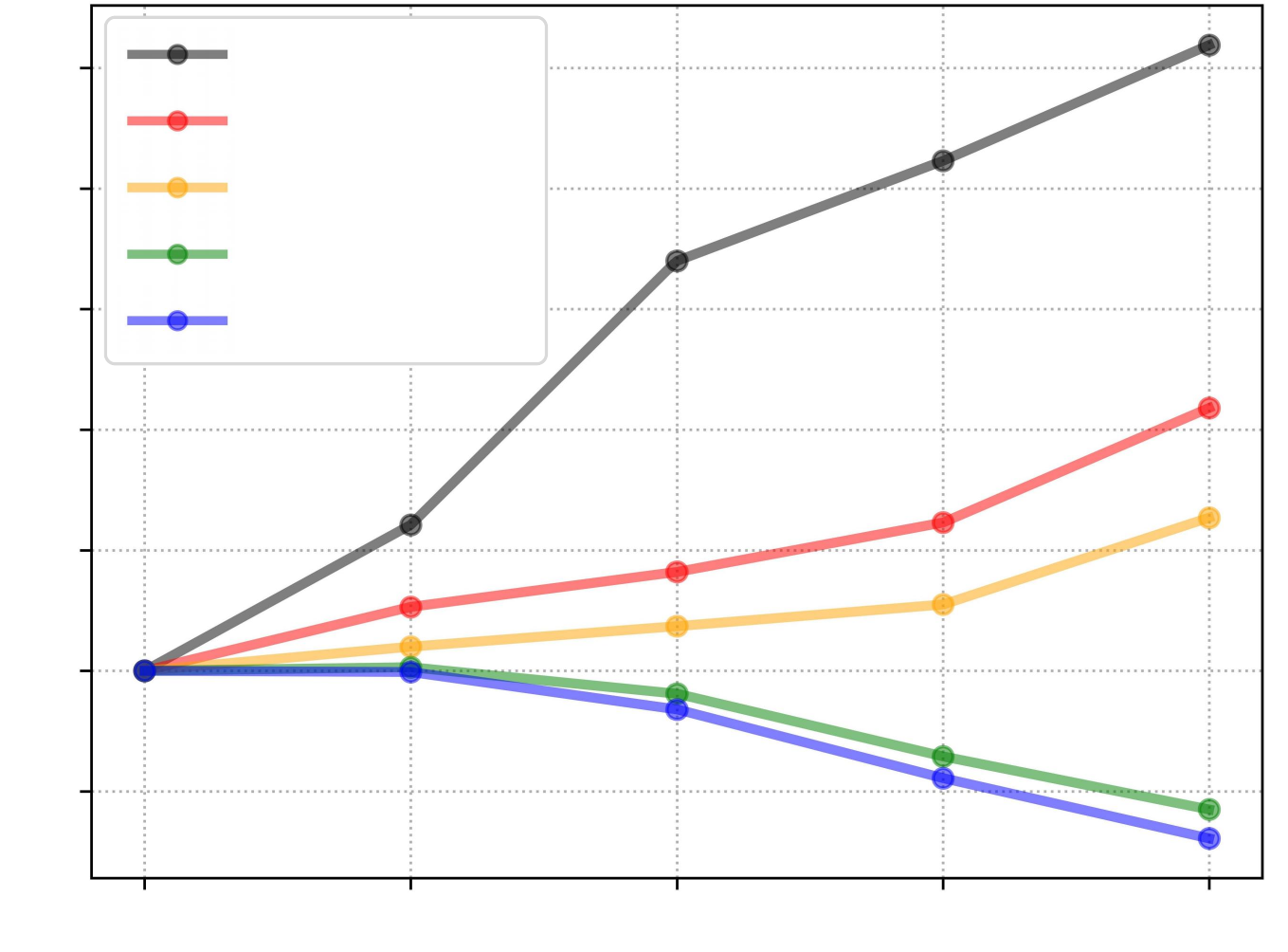}
		\scriptsize
		\put(18,68.7){$z^{out},\beta=1$}
  	\put(18,63.4){baseline}
  	\put(18,58.5){$z^{in},\beta=0.1$}
        \put(18,53.4){$z^{in},\beta=0.3$}
        \put(18,48.2){$z^{in},\beta=0.5$}
        \put(0.4,11.2){-10}
		\put(3.5,20.4){0}
		\put(1.5,30){10}	
		\put(1.5,39.4){20}
		\put(1.5,48.7){30}
		\put(1.5,58){40}
        \put(1.5,67.5){50}
		\put(6.5,1){ZP$^{0,1}$}
		\put(27,1){ZP$^{1,2}$}
		\put(47.8,1){ZP$^{2,3}$}
		\put(68.4,1){ZP$^{3,4}$}
		\put(89,1){ZP$^{4,5}$}
		\end{overpic}
	\caption{ZP relative to ZP$^{0,1}$. The zone performance can be further extremely centralized if the supervision signal strength is reduced in the border zone, and reversely, be anti-centralized if the supervision signal strength is reduced in the central zone.}
  \label{fig:control-zone-performance}
\end{figure}
We visualize the relative ZP in \figref{fig:control-zone-performance}, where all the ZPs are subtracted by ZP$^{0,1}$.
One can see that the centralized spatial bias can be further aggravated if we weaken the supervision signals by reducing the number of positive samples in the border zone.
Reversely, we can even achieve an anti-centralized spatial bias if we weaken the supervision signals in the central zone.
This indicates that the supervision signal strength does have an impact on the zone performance.
Given the above analysis, we finally discuss one possible solution for addressing the spatial disequilibrium problem under the annular zone partition.

\subsection{Spatial equilibrium learning}\label{sec:5.2}

Most existing research on object detection focuses on pursuing higher detection performance at the image level while neglecting optimization at the zone level, resulting in serious spatial disequilibrium issues for detectors.
In this subsection, we introduce a possible solution, termed Spatial Equilibrium Learning as the beginning of relieving the spatial disequilibrium problem.
We first introduce the spatial weight, which maps the anchor point coordinates $(x^{a},y^{a})$ to a scalar $\alpha(x^{a},y^{a})$ by
\begin{equation} \label{eq:spatial_weight}
    \alpha(x,y) = 2\max\left\{||x-\frac{W}{2}||_1\frac{1}{W}, ||y-\frac{H}{2}||_1\frac{1}{H}\right\} \in [0,1],
\end{equation}
where $W$ and $H$ are the width and the height of the image.
The spatial weight can be easily plugged into the existing detection pipelines with few modifications.
The principle is simple and multi-optional.
Here, we offer two implementations as follows.

\begin{figure*}[!t]

	\centering
    \small
	\setlength{\tabcolsep}{1pt}
	\setlength{\abovecaptionskip}{3pt}
    \begin{overpic}[width=1\textwidth]{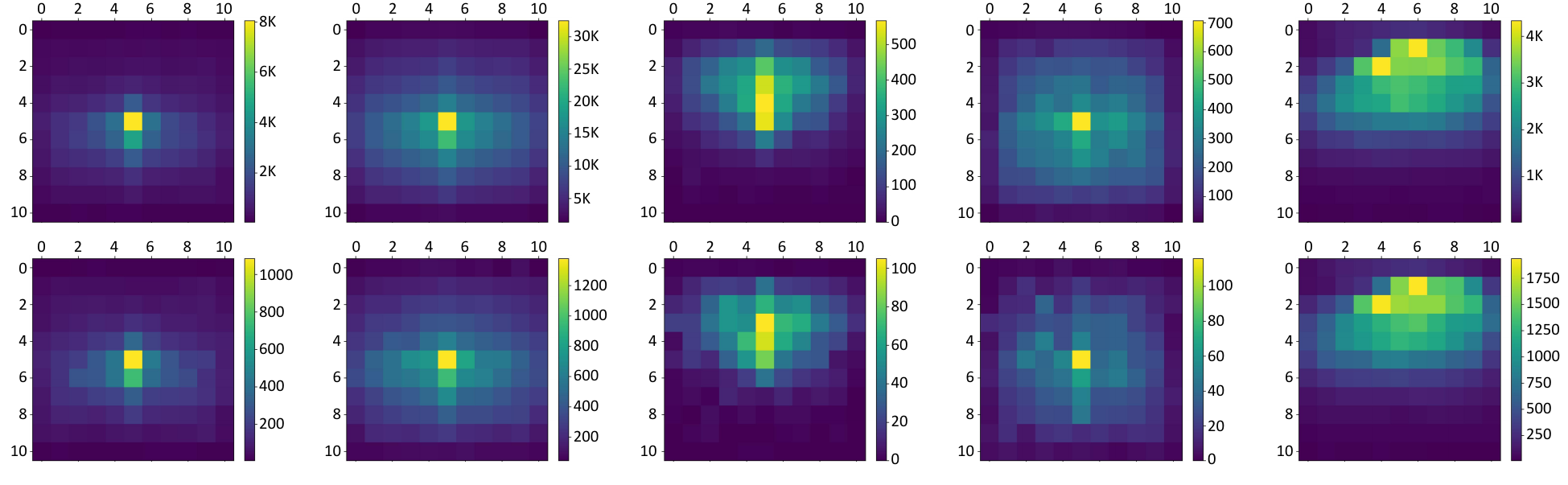}
    \scriptsize
		\put(4,-0.5){PASCAL VOC}
		\put(25.4,-0.5){MS COCO}
		\put(45.9,-0.5){Face Mask}
		\put(67.6,-0.5){Fruit}
		\put(87.2,-0.5){Helmet}
	\end{overpic}
    \vspace{-5pt}
	\caption{The photographer's bias in the 5 object detection datasets. We count the center points for all the ground-truth boxes. The images are divided into $11\times11$ zones. First row: Train set. Second row: Test set.
		}
	\label{fig:5dataset}
\end{figure*}
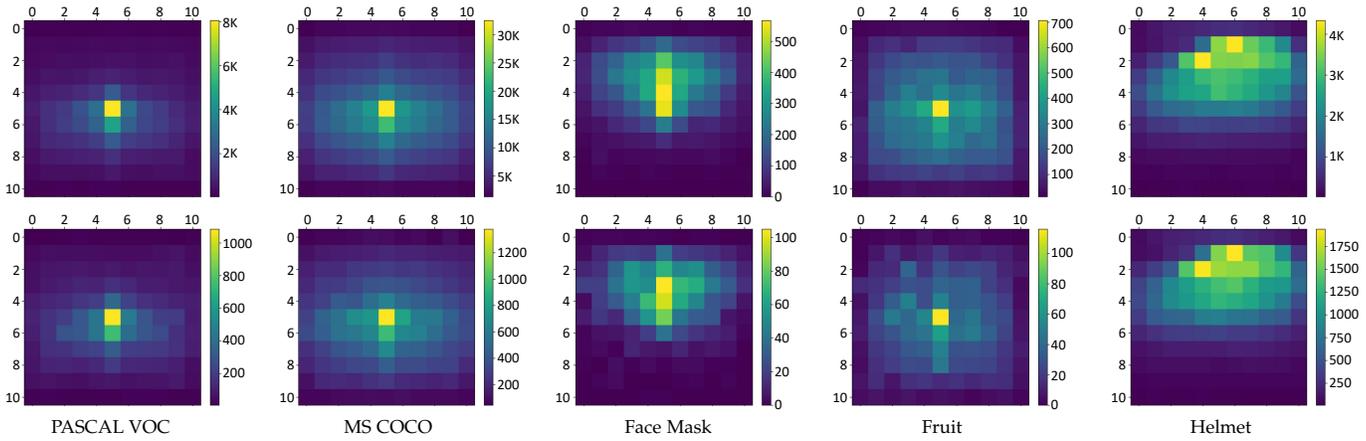
\myPara{1) Spatial equilibrium label assignment (SELA).}
In this approach, the key idea is to consider the spatial weight as an additional constraint term when making the criterion rule for label assignment.
Since most of the label assignment algorithms have their own sophisticated implementations, in the following, we provide a specific application description of the classic ATSS \cite{ATSS}, simply because of its brevity.
Given the positive IoU threshold $t$, which is calculated by considering the statistical characteristics of the objects.
The ATSS criterion follows the same rule as the max-IoU assignment \cite{lin2017focal,yolov3,fasterrcnn}, i.e., $\textrm{IoU}(\bm{B}^{a},\bm{B}^{gt})\geqslant t$,
where $\bm{B}^{a}$ and $\bm{B}^{gt}$ denote the preset anchor boxes and the ground-truth boxes.
The SELA process is represented as:
\begin{equation}\label{eq:SELA}
    \textrm{IoU}(\bm{B}^a,\bm{B}^{gt})\geqslant t-\gamma\alpha(x^{a},y^{a}),
\end{equation}
where $\gamma\geqslant0$ is a hyperparameter.
It can be seen that SELA relaxes the positive sample selection conditions for objects near the image borders.
Therefore, more anchor points will be selected as positive samples for them.
Notice that the above application is actually a frequency-based approach, just like many of the class rebalance sampling strategies proposed for the long-tail class imbalance problem \cite{Kang2020Decoupling,mahajan2018exploring}.

\myPara{2) Spatial equilibrium loss (SE loss).}
In this approach, we adopt the cost-sensitive learning method.
We take the spatial weight term $1+\gamma\alpha(x^{a},y^{a})$ as an additional weight factor for the classification and the bounding box regression losses.
As such, a larger gradient flow will be generated in the border zone so that the network pays more attention to the border objects.

\myPara{Future directions:}
There are more potentially promising solutions toward spatial equilibrium that warrant future study.
For example, designing an appropriate data augmentation \cite{kim2020m2m,chou2020remix,wang2019implicit}, more specifically, increasing data augmentation to make up for the insufficient sampling frequency for the objects near the image borders, might be a promising solution.
Besides, since the spatial disequilibrium problem is formally equivalent to the class imbalance problem, some improved re-balancing methods may also bring gains to spatial equilibrium learning, e.g., class-balanced loss \cite{cui2019class}, transfer learning \cite{wang2017learning,chu2020feature}, and representation learning \cite{huang2016learning,dong2017class,liu2019large}, etc.
There is also a very exciting area for future work in pursuing more solutions to address the two problems in a unified form.
Furthermore, our approaches mainly consider the annular zone partition, which is designed to balance the detection ability between the central zone and the border zone.
For some special applications that care more about some regions of interest, the design ethos can be versatile and contingent on the practical application.

\section{Quantitative Evaluation} \label{sec:results}
In this section, we conduct comprehensive evaluations on 10 popular object detectors and 5 object detection datasets.

\subsection{Experimental setups}\label{sec:6.1}
\myPara{Detectors and metrics.}
All the object detectors we evaluate can be downloaded from MMDetection \cite{mmdetection} or their official websites.
We follow the standard Average Precision evaluation protocol.
To comprehensively evaluate the detectors, various metrics are reported, including 5 ZPs, the variance of 5 ZPs, and the traditional metric AP.

\myPara{Datasets.}
All the datasets we used are publicly available and can be downloaded from their official websites or Kaggle.
The object distributions of the 5 datasets can be seen in \figref{fig:5dataset}.

\textit{PASCAL VOC \cite{voc}}
is one of the most widely used object detection benchmark under natural scenes, which contains 20 classes.
We adopt the classical 07+12 training and testing protocol, i.e., the train set contains the union of VOC 2007 trainval and VOC 2012 trainval (totally 16551 images) and the test set contains VOC 2007 test (4952 images).

\textit{MS COCO \cite{coco}}
is another recently popular benchmark with much larger scale, containing 80 classes under natural scenes.
We adopt COCO 2017 train (118K images) for training and COCO 2017 val (5K images) for evaluation.

\textit{Face mask detection \cite{facemask}}.
With COVID-19 raging around the world, face mask detection is a widespread and necessary visual application.
The dataset consists of 5,865 images for training and 1,035 images for testing. 
There are 2 classes. One is face and the other is mask.

\textit{Fruit detection \cite{fruit}}
is widely used in industrial assembly line sorting and commodity classification.
The dataset consists of 3,836 train images and 639 test images.
11 common fruits are included, e.g., apple, grape, and lemon, etc.

\textit{Helmet detection \cite{helmet}}
is a safety vision application that is often used on construction sites to detect whether workers and visitors are wearing helmets.
It contains 15,887 images for training and 6,902 images for testing.
Two classes, head and helmet, are used.

\myPara{Setups of spatial equilibrium learning.}
For spatial equilibrium learning evaluation, the implementation is based on the MMDetection \cite{mmdetection} framework, and the ablation study is conducted on GFocal \cite{gfocal} with ResNet \cite{ResNet} backbone and FPN \cite{FPN} neck.
We use ResNet-18 for VOC 07+12 and 3 application datasets, and adopt ResNet-50 for MS COCO.
The learning rate is linearly scaled by the linear scaling rule \cite{goyal2017accurate} according to the number of GPUs. 
The training epochs are set to 12 for all the experiments.
We set $\gamma=0.2$ in \eqref{eq:SELA}, and all the other hyper-parameters remain unchanged for a fair comparison.

\begin{table}[!t]
  \centering
  \scriptsize
  \setlength{\tabcolsep}{2.5pt}
  \caption{Zone evaluation on the existing popular object detectors. 
    5 zone precisions (ZP), the variance of ZP, and the traditional metric AP are reported. 
    The results are reported on COCO 2017 val.
    ``Cas.'': Cascade.
    \textbf{R}: ResNet \cite{ResNet}. 
    \textbf{X}: ResNeXt-32x4d \cite{xie2017aggregated}. 
    \textbf{PVT-s}: Pyramid vision transformer-small \cite{PVT}.
    \textbf{CNeXt-T}: ConvNext-T \cite{convnet}.
  }
  \begin{tabular}{lccccccc}
	Detector & AP  & Var. & ZP$^{0,1}$ & ZP$^{1,2}$ & ZP$^{2,3}$ & ZP$^{3,4}$ & ZP$^{4,5}$ \\ \hline
    DETR (\textbf{R}-50) \cite{DETR} 
     & 40.1 & 26.9 & 29.8 & 36.2 & 39.8 & 39.1 & 45.7 \\
    RetinaNet (\textbf{PVT-s}) \cite{PVT} 
     & 40.4 & 19.7 & 30.8 & 36.9 & 39.0 & 37.4 & 44.6 \\
    Cascade R-CNN (\textbf{R}-50) \cite{cascadercnn} 
     & 40.3 & 18.7 & 30.9 & 36.6 & 39.2 & 38.6 & 44.2 \\
    GFocal (\textbf{R}-50) \cite{gfocal} 
    & 40.1 & 16.9 & 31.1 & 37.5 & 39.4 & 38.5 & 43.8 \\\hline
    Cas. Mask R-CNN (\textbf{R}-101) \cite{cascadercnn} 
     & 45.4 & 22.4 & 34.7 & 41.6 & 44.3 & 44.4 & 49.1 \\
    Sparse R-CNN (\textbf{R}-50) \cite{sparsercnn} 
     & 45.0 & 21.6 & 35.8 & 41.9 & 43.4 & 44.0 & 50.3 \\
    YOLOv5-m \cite{yolov5} 
     & 45.2 & 12.9 & 36.0 & 42.3 & 44.5 & 43.2 & 46.7 \\\hline
    Deform. DETR (\textbf{R}-50) \cite{deformabledetr} 
     & 46.1 & 23.2 & 36.3 & 42.6 & 45.6 & 45.1 & 51.2 \\
    Sparse R-CNN (\textbf{R}-101) \cite{sparsercnn} 
     & 46.2 & 21.2 & 36.9 & 42.9 & 44.9 & 44.7 & 51.3 \\
    Cas. Mask R-CNN (\textbf{X}-101) \cite{cascadercnn} 
    & 46.1 & 21.1 & 36.1 & 42.0 & 44.8 & 45.9 & 49.9 \\
    Mask R-CNN (\textbf{CNeXt-T}) \cite{convnet} 
    & 46.2 & 17.6 & 36.7 & 41.9 & 44.5 & 43.6 & 49.7 \\
    GFocal (\textbf{X}-101) \cite{gfocal} 
    & 46.1 & 15.7 & 37.0 & 43.5 & 45.0 & 44.4 & 49.3 \\
    VFNet (\textbf{R}-101) \cite{VFNet} 
     & 46.2 & 15.6 & 36.7 & 43.0 & 45.0 & 44.5 & 48.8 \\ 
  \end{tabular}
  \label{tab:detectors}
  \vspace{-5pt}
\end{table}

\subsection{Zone evaluation on various object detectors}\label{sec:6.2}

Although traditional evaluation provides good guidance for the overall performance of detectors, little is known about the spatial bias of detectors, as well as the location and degree.
Here, we select various object detectors that have different detection pipelines but with the same level of traditional metrics.
They are popular, representative, and considered to be the milestones in modern object detection: one-stage dense detectors (RetinaNet \cite{lin2017focal}, GFocal \cite{gfocal}, VFNet \cite{VFNet}, YOLOv5 \cite{yolov5}), multi-stage dense-to-sparse detectors (R-CNN series \cite{fasterrcnn,cascadercnn,maskrcnn}), and sparse detectors (DETR series \cite{DETR,deformabledetr} and Sparse R-CNN \cite{sparsercnn}).
The quantitative results are reported in Table \ref{tab:detectors}.

There are several interesting observations:

1) \textit{Spatial bias is quite common}.
One can see that all the detectors show a clear centralized zone performance, i.e., performing well in the central zones ($z^{3,4}, z^{4,5}$) but unsatisfactorily in the border zones ($z^{0,1},z^{1,2}$).
This confirms the existence and universality of spatial bias, and we successfully quantify the defects of object detectors shown in \figref{fig:visual} for the first time.

2) \textit{Their spatial equilibria vary significantly.}
There is a large gap of 12.9 to 26.9 in ZP variance.
Particularly, we find that the sparse detectors, e.g., DETR series and Sparse R-CNN, tend to produce a large ZP variance, while the one-stage dense object detectors perform better in spatial equilibrium (lower ZP variance).
It shows that the DETR-based detectors do not align with CNN-based detectors in terms of spatial equilibrium.
We conjecture that it may be attributed to the global information captured by the self-attention mechanism.
The sparse detectors first extract features by CNNs and then process the features by a series of attention modules.
The training is more dynamic, and we hypothesize that the central objects may receive more attention.
Clearly, one can conclude that there must be some factors that lead to different spatial equilibria among detectors, including but not limited to the neural network architecture designs, optimization, and training strategies.
Yet currently, we are unaware of which components or algorithm designs have an impact on spatial bias.
We believe that further research on this topic in the future will be interesting, and the study is of great possibility to find the key to addressing the spatial disequilibrium problem.

\begin{table}[!t]
  \footnotesize
  \centering
  \setlength{\tabcolsep}{3pt}
  \caption{Zone evaluation on VOC 07+12, COCO 2017, and 3 application datasets, i.e., face mask detection, fruit detection, and helmet detection. 5 zone precisions (ZP), the variance of ZP, and the traditional metric AP are reported. The results are reported on GFocal \cite{gfocal}.
  }\vspace{-0pt}
  \begin{tabular}{lccccccc}
	Dataset & AP  & Var. & ZP$^{0,1}$ & ZP$^{1,2}$ & ZP$^{2,3}$ & ZP$^{3,4}$ & ZP$^{4,5}$ \\ \hline
    VOC 07+12 
     & 52.2 & 53.6 & 34.3 & 39.6 & 42.5 & 46.6 & 56.1 \\
    COCO 2017 
    & 40.1 & 16.9 & 31.1 & 37.5 & 39.4 & 38.5 & 43.8 \\
    Face Mask 
    & 71.3 & 13.1 & 60.4 & 67.1 & 69.0 & 68.8 & 70.9 \\
    Fruit 
    & 76.6 & 56.2 & 60.8 & 69.9 & 71.2 & 75.3 & 83.8 \\
    Helmet
    & 49.7 & 3.0 & 45.9 & 47.9 & 50.3 & 50.6 & 47.8 \\
  \end{tabular}
  \label{tab:datasets}
\end{table}

\begin{figure}[!t]
  \vspace{-5pt}
	\centering
            \begin{overpic}[width=0.48\textwidth]{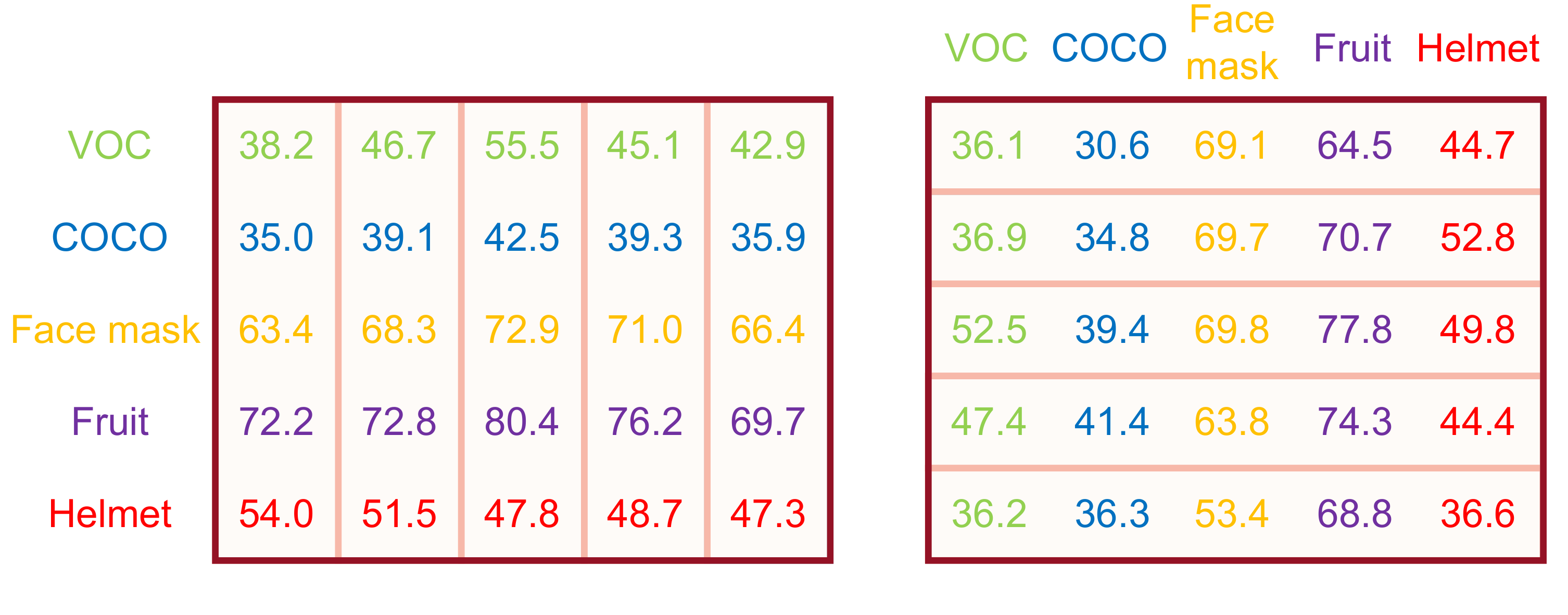}
            \small
            \put(14,-0.5){(a) 5 zones along x-axis}
            \put(60,-0.5){(b) 5 zones along y-axis}
            \end{overpic}
	\caption{Two designs of zone partition. ZP is reported.
		}
  \label{fig:otherzonepartition}
   \vspace{-5pt}
\end{figure}

3) \textit{Traditional evaluation fails to capture spatial bias.}
It can be seen that GFocal (R-50) and DETR (R-50) achieve the same AP score of 40.1.
However, the traditional metrics provide nothing about how much the detection performance is in a zone.
Our zone evaluation shows that GFocal performs better in the border zones $z^{0,1}$ and $z^{1,2}$, while DETR performs better in the zones $z^{2,3}$, $z^{3,4}$, and $z^{4,5}$.
Similarly, Deformable DETR (\textbf{R}-50) \cite{deformabledetr} achieves the same traditional AP as GFocal (\textbf{X}-101).
The zone evaluation shows that Deformable DETR performs significantly better than GFocal in the central zones $z^{3,4}$, $z^{4,5}$,
whereas worse in the border zones $z^{0,1}$, $z^{1,2}$.
And these performance discrepancies are concealed by the traditional evaluation.
In addition, it is interesting to see that the AP metrics are exactly between ZP$^{3,4}$ and ZP$^{4,5}$, which indicates that the detection performance in 96\% of the image area is actually lower than AP.

\textit{Implications}: The above results reveal the performance characteristics of the detectors, which help us better understand the behavior of the object detectors, and encourage us to rethink the choice of detectors when deploying to the application scenarios.
Furthermore, it is also worth studying which components in the detection pipelines lead to such performance discrepancies, e.g., self-attention mechanism, label assignment, etc.

\subsection{Zone evaluation on various datasets}\label{sec:6.3}

Table \ref{tab:datasets} reports the quantitative detection results on PASCAL VOC 07+12, MS COCO val2017, and 3 application datasets.
We have the following observations:

1) One can see that the detection performance varies across the zones.
The zone closest to the image border, i.e., $z^{0,1}$, has consistently the lowest detection performance.
In contrast, the central zone $z^{4,5}$ has the highest performance in almost all of these cases.

2) There is also a representative case, i.e., the helmet dataset, having only 3.0 ZP variance.
This indicates that the helmet dataset achieves the best spatial equilibrium under the scenario of annular zone partition, whereas the other datasets have a clear spatial disequilibrium problem.
For example, the variance of ZP is 53.6 on PASCAL VOC and 56.2 on the fruit dataset.

3) If we switch to other zone partitions, e.g., 5 strip zones along x-axis and 5 strip zones along y-axis (see \figref{fig:otherzonepartition}(a)(b)), their spatial equilibria change.
In Table \ref{tab:ZPvar}, the face mask and helmet datasets increase ZP variance to 39.6 and 30.5 in the scenario of 5 zones along y-axis, respectively, while the fruit dataset decreases the ZP variance significantly in both cases.

\begin{table}[!t]
  \footnotesize
  \centering
  \setlength{\tabcolsep}{3pt}
  \caption{ZP variance on three types of zone partitions. Spatial bias is an external manifestation of detectors, whereas spatial equilibrium corresponds to a given zone partition.}
  \begin{tabular}{ccccccc}
    Zone partition     & VOC  & COCO & Face Mask & Fruit & Helmet \\ \hline
    5 Annular zones & 53.6  & 16.9 & 13.1 & 56.2 & 3.0 \\
    5 zones along x-axis    & 32.3 & 7.2 & 11.2 & 13.7 & 6.4\\
    5 zones along y-axis  & 46.7 & 14.0 & 39.6 & 20.8 & 30.5\\
  \end{tabular}
  \label{tab:ZPvar}
\end{table}

\begin{table}[!t]
  \centering
  \footnotesize
  \setlength{\tabcolsep}{5pt}
  \caption{Evaluation for hyper-parameter $\gamma$ in SELA. 
    5 zone precisions (ZP), the variance of ZPs, and the traditional metric AP are reported. 
    $\gamma=0$ denotes the baseline GFocal.
    Lower variance, better spatial equilibrium. (Dataset: VOC 07+12)
  }
  \begin{tabular}{c|ccccccc}
    $\gamma$ & AP & Var. & ZP$^{0,1}$ & ZP$^{1,2}$ & ZP$^{2,3}$ & ZP$^{3,4}$ & ZP$^{4,5}$ \\ \hline
	0  & 52.2 & 53.6 & 34.3 & 39.6 & 42.5 & 46.6 & \textbf{56.1} \\ \hline
    0.1 & 52.5 & 44.5 & 35.9 & 40.6 & 42.1 & 46.6 & 55.6 \\
    0.2 & \textbf{52.8} & \textbf{37.7} & \textbf{37.6} & 40.3 & \textbf{43.8} & \textbf{46.9} & 55.4 \\
    0.3 & \textbf{52.8} & 37.3 & 37.4 & \textbf{41.5} & 43.6 & \textbf{46.9} & 55.6 \\
    0.4 & 52.0 & 46.3 & 35.0  & 38.9 & 42.6 & 46.6 & 54.8 \\
  \end{tabular}
  \label{tab:gamma}
  \vspace{-5pt}
\end{table}

\textit{Implications}: 
The zone evaluation provides a fresh perspective that sheds light on the limitations of object detectors.
It can be seen that spatial bias is also a natural characteristic of object detectors and they are difficult to perform perfect spatial equilibrium for an arbitrary zone partition.
The above results indicate that the ZP variance is a set function related to zone partition.
The annular zone partition mainly considers the balance of detection ability between the inner zone and the outer one, which is a good choice in practice since the centralized photographer's bias is ubiquitous in visual datasets.
Nevertheless, it should be noted that the zone partition is flexible and is able to be customized to any shape based on the application scenarios.

\subsection{Evaluation on spatial equilibrium learning}\label{sec:6.4}

Finally, we provide the evaluation of spatial equilibrium learning.
The ablation studies are conducted by using GFocal and we adopt the first approach, i.e., spatial equilibrium label assignment (SELA), by default.

\myPara{Hyperparameter $\gamma$.}
Recall that the implementation of SELA only involves one hyper-parameter $\gamma$ in \eqref{eq:SELA}.
$\gamma$ controls the magnitude of the spatial weight.
A larger $\gamma$ increases more positive samples for objects near the image borders. 
As shown in Table \ref{tab:gamma}, we observe that our SELA can achieve a consistent spatial equilibrium improvement (lower variance) for all the options of $\gamma$.
Too large $\gamma$, e.g. 0.4, will increase much more positive samples for all the zones, leading to a performance drop.
Thus, we set $\gamma$ to 0.2 for PASCAL VOC.
One can see that our SELA can significantly improve the detection performance for the outer zones, e.g., ZP$^{0,1}$, ZP$^{1,2}$, ZP$^{2,3}$ and ZP$^{3,4}$.
As shown in \figref{fig:APcurve-PCC-SCC}(a), while the central zone $z^{4,5}$ has a slight performance drop, the ZP improvement is remarkable in the border zone, which occupies 96\% of the total image area.
This is particularly important for the safety applications in surveillance systems and self-driving cars, as objects may appear anywhere.
The performance of the border zone plays a significant role in robustness detection.
In practice, we set $\gamma=0.1$ for all the other datasets, but it should be noted that there might be a better $\gamma$ for different application scenarios.

\begin{table}[!tb]
  \centering
  \footnotesize
  \setlength{\tabcolsep}{4pt}
  \caption{Analysis of spatial weight. 
    5 zone precisions (ZP), the variance of ZPs, and the traditional metric AP are reported. 
    $\gamma=0.2$. 
    Lower variance, better spatial equilibrium. (Dataset: VOC 07+12)
  }
  \begin{tabular}{c|ccccccc}
    weight & AP & Var. & ZP$^{0,1}$ & ZP$^{1,2}$ & ZP$^{2,3}$ & ZP$^{3,4}$ & ZP$^{4,5}$ \\ \hline
	0  & 52.2 & 53.6 & 34.3 & 39.6 & 42.5 & 46.6 & 56.1 \\
    1 & \textbf{52.8} & 48.3 & 35.7 & 40.2 & 43.3 & \textbf{47.1} & \textbf{56.2} \\
    $\alpha(x^a,y^a)$ & \textbf{52.8} & \textbf{37.7} & \textbf{37.6} & \textbf{40.3} & \textbf{43.8} & 46.9 & 55.4 \\
\end{tabular}
\label{tab:spatial-weight}
\end{table}

\begin{table}[!t]
  \vspace{-3pt}
  \footnotesize
  \centering
  \setlength{\tabcolsep}{3pt}
  \caption{Zone evaluation on PASCAL VOC 07+12, MS COCO 2017, and 3 application datasets, including face mask detection, fruit detection, and helmet detection. 5 zone precisions (ZP), the variance of ZPs, and the traditional metric AP are reported. The detector is GFocal \cite{gfocal}.
  }
  \begin{tabular}{lcccccccc}
	Dataset & SELA & AP  & Var. & ZP$^{0,1}$ & ZP$^{1,2}$ & ZP$^{2,3}$ & ZP$^{3,4}$ & ZP$^{4,5}$ \\ \hline
    \multirow{2}{*}{VOC 07+12}  &
     & 52.2 & 53.6 & 34.3 & 39.6 & 42.5 & 46.6 & 56.1 \\
     & \checkmark & 52.8 & 37.7 & 37.6 & 40.3 & 43.8 & 46.9 & 55.4 \\ \hline
    \multirow{2}{*}{COCO 2017} &
    & 40.1 & 16.9 & 31.1 & 37.5 & 39.4 & 38.5 & 43.8 \\
    & \checkmark & 40.3 & 14.4 & 31.2 & 37.7 & 39.5 & 38.3 & 42.9 \\ \hline
    \multirow{2}{*}{Face Mask} & 
    & 71.3 & 13.1 & 60.4 & 67.1 & 69.0 & 68.8 & 70.9 \\
    & \checkmark & 71.6 & 12.1 & 60.6 & 68.0 & 69.5 & 69.3 & 69.8 \\ \hline
    \multirow{2}{*}{Fruit} &
    & 76.6 & 56.2 & 60.8 & 69.9 & 71.2 & 75.3 & 83.8 \\
    & \checkmark & 77.0 & 33.6 & 65.7 & 69.8 & 72.0 & 76.2 & 82.7 \\ \hline
    \multirow{2}{*}{Helmet} &
    & 49.7 & 3.0 & 45.9 & 47.9 & 50.3 & 50.6 & 47.8 \\
    & \checkmark & 49.9 & 3.1 & 45.9 & 48.5 & 50.5 & 50.6 & 47.9 \\
  \end{tabular}
  \label{tab:datasets-SELA}
  \vspace{-5pt}
\end{table}

\myPara{Spatial weight.}
One may wonder how the performance would go if we directly loose the selection condition for the positive samples without considering their spatial positions.
Here, we conduct an experiment to investigate the effect of the spatial weight.
The quantitative results are reported in Table \ref{tab:spatial-weight}.
If the spatial weight is set to a constant of 1, it means that we directly lower the positive IoU threshold $t$ by
$\textrm{IoU}(\bm{B}^a,\bm{B}^{gt})\geqslant t-\gamma$,
and more positive samples will be selected without spatial discrimination.
One can see that although the performance is boosted, the variance of the 5 ZPs is large.
This indicates that subtracting a constant from the positive IoU threshold cannot change the sampling frequency much, because more positive samples are generated in the central zone.
In contrast, our SELA can significantly reduce the variance, and achieve a much better spatial equilibrium.

\myPara{SELA on various datasets.}
Table \ref{tab:datasets-SELA} shows us promising results that our SELA can achieve a better spatial equilibrium for object detection.
In particular, we reduce the variance by a large margin in terms of ZP.
For example, we successfully lower the variance of ZP by -15.9, -2.5, -1.0, and -22.6 on PASCAL VOC, MS COCO, and face mask/fruit detection.
It demonstrates that our SELA can improve the spatial equilibrium for multiple application scenarios without the sacrifice of AP.

\begin{table}[!t]
  \footnotesize
  \centering
  \setlength{\tabcolsep}{2.1pt}
  \caption{Evaluation of SELA with various backbone networks. 
    5 zone precisions (ZP), the variance of ZPs, and the traditional metric AP are reported. \textbf{X}: ResNeXt \cite{xie2017aggregated}. (Dataset: VOC 07+12)
  }
  \begin{tabular}{ccccccccc}
    Model & SELA & AP & Var. & ZP$^{0,1}$ & ZP$^{1,2}$ & ZP$^{2,3}$ & ZP$^{3,4}$ & ZP$^{4,5}$ \\ \hline
	\multirow{2}{*}{ResNet-18} & & 52.2 & 53.6 & 34.3 & 39.6 & 42.5 & 46.6 & 56.1 \\
    & \checkmark & 52.8 & 37.7 & 37.6 & 40.3 & 43.8 & 46.9 & 55.4 \\ \hline
	\multirow{2}{*}{ResNet-50} &  & 56.1 & 41.5 & 40.9 & 44.6 & 46.7 & 51.0 & 59.7 \\
    & \checkmark & 56.2 & 32.2 & 43.3 & 44.6 & 47.3 & 50.4 & 59.2 \\ \hline
	\multirow{2}{*}{\textbf{X}-101-32x4d-DCN} &  & 64.0 & 37.1 & 48.7 & 53.1 & 55.0 & 58.0 & 66.9 \\
    & \checkmark & 64.3 & 31.0 & 50.2 & 54.1 & 55.9 & 57.7 & 66.9 \\
  \end{tabular}
  \label{tab:backbone}
\end{table}

\begin{table}[!t]
\vspace{-5pt}
  \footnotesize
  \centering
  \setlength{\tabcolsep}{3pt}
  \caption{Generality of spatial equilibrium learning. The cost-sensitive learning based method, SE loss, is adopted. 5 zone precisions (ZP), the variance of ZPs, and the traditional metric AP are reported. (Dataset: VOC 07+12)
  } \vspace{-5pt}
  \begin{tabular}{lcccccccc}
	Detector & SE loss & AP  & Var. & ZP$^{0,1}$ & ZP$^{1,2}$ & ZP$^{2,3}$ & ZP$^{3,4}$ & ZP$^{4,5}$ \\ \hline
    \multirow{2}{*}{GFocal \cite{gfocal}} &
     & 52.2 & 53.6 & 34.3 & 39.6 & 42.5 & 46.6 & 56.1 \\
     & \checkmark & 52.5 & 41.6 & 37.1 & 40.6 & 42.9 & 46.5 & 56.0 \\ \hline
    \multirow{2}{*}{DW \cite{DW}} &
    & 51.8 & 32.6 & 38.4 & 39.9 & 43.3 & 45.7 & 54.6 \\
    & \checkmark & 52.7 & 25.9 & 39.8 & 41.2 & 44.4 & 46.8 & 54.2 \\ \hline
    \multirow{2}{*}{DDOD \cite{DDOD2021}} & 
    & 51.1 & 22.6 & 38.4 & 40.0 & 42.2 & 45.2 & 51.9 \\
    & \checkmark & 51.5 & 20.8 & 40.9 & 40.1 & 42.6 & 45.8 & 52.7 \\ \hline
    \multirow{2}{*}{DINO \cite{DINO}} & 
    & 61.5 & 47.6 & 47.1 & 48.4 & 53.0 & 57.1 & 66.2 \\
    & \checkmark & 61.7 & 46.7 & 47.4 & 48.5 & 53.4 & 57.1 & 66.3 \\
  \end{tabular}
  \label{tab:other-detectors}
  \vspace{-10pt}
\end{table}

\begin{figure*}[!t]
    \centering
    \small
    \setlength{\tabcolsep}{1pt}
    \setlength{\abovecaptionskip}{3pt}
    \begin{overpic}[width=1\textwidth]{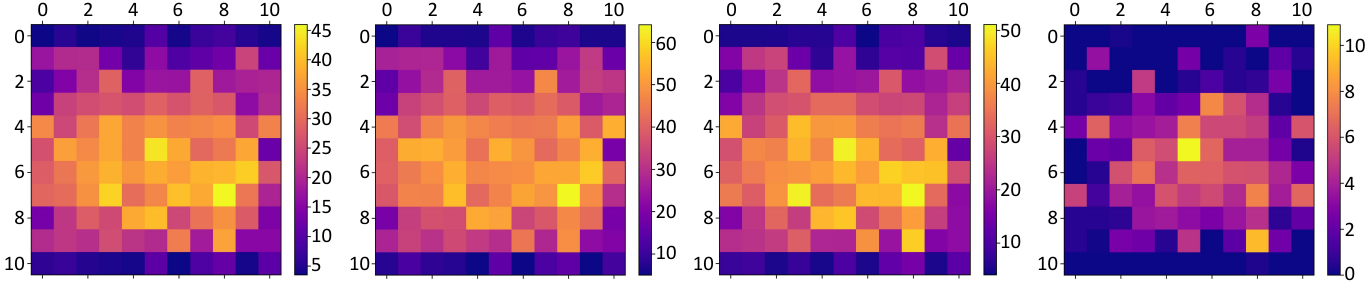}
    \put(10.2,0.2){ZP}
    \put(35,0.2){ZP$_{50}$}
    \put(60.2,0.2){ZP$_{75}$}
    \put(85.4,0.2){ZP$_{95}$}
    \end{overpic}
    \vspace{-10pt}
    \caption{Zone evaluation on $11\times 11$ square zones. The model is GFocal. The results are reported on VOC 07+12.
    }
    \label{fig:121ZP}
\end{figure*}

\begin{figure*}[!t]
    \centering
    \small
    \setlength{\tabcolsep}{1pt}
    \setlength{\abovecaptionskip}{3pt}
    \begin{overpic}[width=0.98\textwidth]{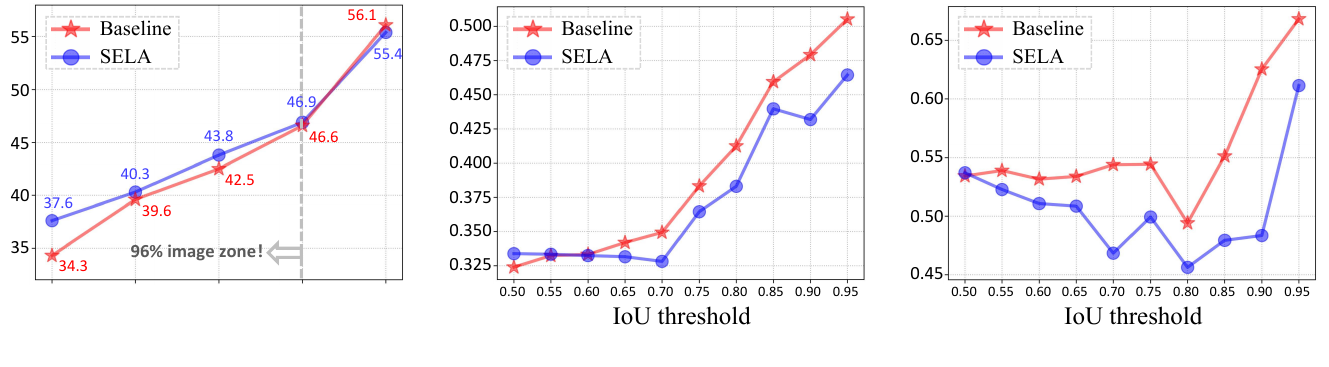}
    \small
    \put(2.1,4.2){ZP$^{0,1}$}
    \put(8.3,4.2){ZP$^{1,2}$}
    \put(14.6,4.2){ZP$^{2,3}$}
    \put(20.8,4.2){ZP$^{3,4}$}
    \put(27.1,4.2){ZP$^{4,5}$}
    
		\put(8,0.5){(a) Effect of SELA}
		\put(47,0.5){(b) PCC}
		\put(81,0.5){(c) SCC}
	\end{overpic}
    \vspace{-5pt}
    \caption{(a) ZP against the zone. (b) Pearson Correlation Coefficient (PCC) between the mZP and the object distribution (center counts) against the IoU threshold. (c) Spearman Correlation Coefficient (SCC) between the mZP and the object distribution against the IoU threshold. Our SELA can substantially reduce these correlations under most of IoU thresholds, indicating a better spatial equilibrium. The baseline model is GFocal. The results are reported on VOC 07+12.
    }
    \label{fig:APcurve-PCC-SCC}
    \vspace{-5pt}
\end{figure*}
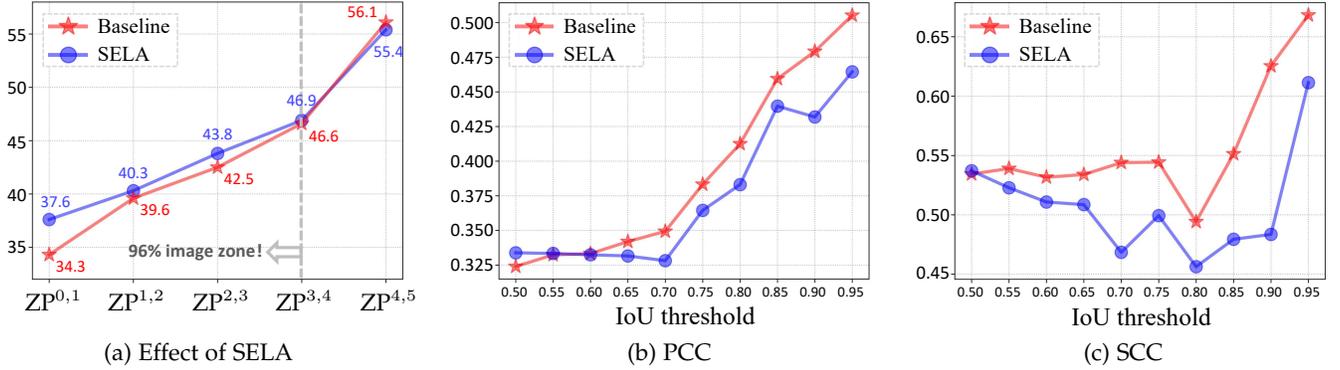

\myPara{Generality of spatial equilibrium learning.}
We further provide more experiments to verify the effectiveness of spatial equilibrium learning on various backbone networks.
Table \ref{tab:backbone} exhibits that our SELA can notably improve the spatial equilibrium for all the 3 backbone networks, i.e., lower variance.
We also conduct experiments to check out the generality of spatial equilibrium learning by incorporating it into 3 more detectors, DW \cite{DW}, DDOD \cite{DDOD2021}, and DETR-like detector DINO \cite{DINO}.
Here, we adopt spatial equilibrium loss (SE loss) and we enlarge the training losses for the objects near the image border.
Table \ref{tab:other-detectors} reports the quantitative results of SE loss for these 4 object detectors.
As shown, our method can substantially reduce the ZP variance for the 4 detectors, indicating that a better spatial equilibrium is achieved.
This shows the generalized ability of our method to improve the spatial robustness of detectors without bells and whistles.

We also note that compared to the CNN-based object detectors, our method produces a slight improvement of spatial equilibrium on DINO.
This may be attributed to the different optimization processes between DETR-like detectors and others.
The number of positive samples is quite limited in DETR-like detectors since they use one-to-one Hungarian matching, while it is much more abundant in dense object detectors as they adopt one-to-many assignments.
Thus, our SE loss is more helpful in alleviating the imbalance of supervision signal strength on dense object detectors.
This implies that improving spatial equilibrium for DETR-like detectors could be more challenging.
We hope this work could inspire more solutions to address the disequilibrium problem of DETR-like detectors in the future.

\myPara{Correlation with object distribution.}
We further provide the correlation between the zone metrics and the object distribution.
We define a finer zone partition, which is the same as the zone partition for counting the centers of objects, i.e., $11\times11$ square zones (see \figref{fig:5dataset}).
Then we evaluate the detection performance in the 121 zones one by one.
We plot the ZP of the 121 zones in Fig.~\ref{fig:121ZP}.
It can be seen that the ZP distribution is similar to the object distribution (Fig.~\ref{fig:5dataset}), i.e., the same centralized trend.
To investigate the correlation between the zone metrics and the object distribution, we further calculate the Pearson Correlation Coefficient (PCC) and the Spearman Correlation Coefficient (SCC) between the mZPs and the object distribution of the test set.
As shown in Fig.~\ref{fig:APcurve-PCC-SCC}(b) and Fig.~\ref{fig:APcurve-PCC-SCC}(c), we get the following deep reflections on the spatial bias.
We first note that all the PCCs $>0.3$ in \figref{fig:APcurve-PCC-SCC}(b), which indicates that the detection performance is moderately linear correlated with the object distribution.
As a reminder, the PCC only reflects the linear correlation of two given vectors, while it may fail when they are curvilinearly correlated.
In \figref{fig:APcurve-PCC-SCC}(c), the Spearman correlation reflects a higher ranking correlation between the mZPs and the object distribution with all the SCCs $>0.45$.
This illustrates that the detection performance has a moderate-to-high correlation with the object distribution.
In general, our SELA substantially reduces these correlations, indicating a lower correlation with object distribution and better spatial equilibrium.

\begin{figure*}[!t]
	\centering
		\includegraphics[width=1\textwidth]{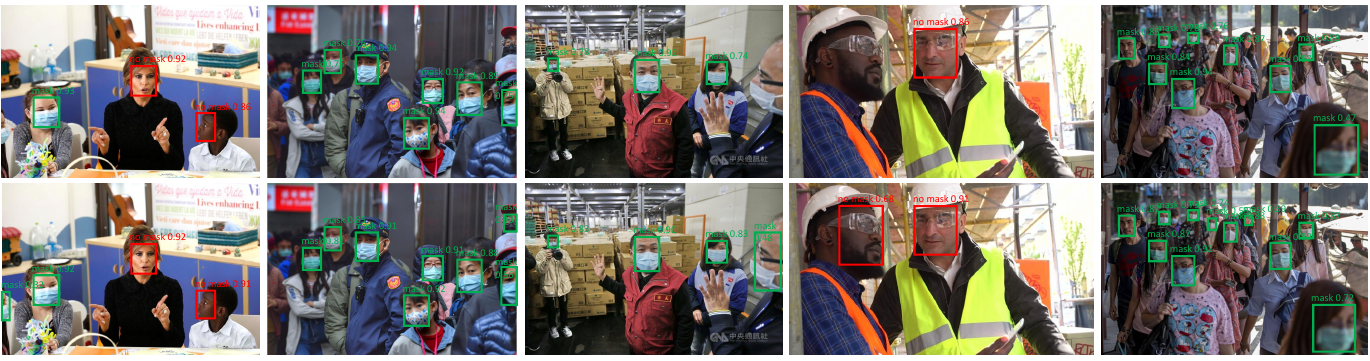}
   \vspace{-10pt}
	\caption{Illustration of detection results for GFocal (first row) and GFocal + SELA (second row). Our method boosts the detection performance for the border zone. Zoom in for a better view.
		}
	\label{fig:base-SELA}
\end{figure*}

\myPara{Visualization of Detection.}
We visualize the detection results of SELA in Fig. \ref{fig:base-SELA}.
Our method can improve the detection performance of the border zone.
We believe that further exploration of spatial equilibrium is clearly worthy and important for robust detection applications.

\subsection{Other attempts toward spatial equilibrium}\label{sec:6.5}
As we discussed in \secref{sec4}, our results show that the object scale and the absolute positions of objects barely influence the spatial bias.
The discrepancy of object data patterns between the zones plays a significant role in spatial bias.
In this subsection, we investigate more possible factors and see how the spatial equilibrium changes.
The first one is the padding operation.
We follow the work \cite{kayhan2020translation} to set the padding way as full-conv, which is translation-invariant as \cite{kayhan2020translation} demonstrated.
We replace all the convolution kernels of the head networks with full-conv.
The second is the effect of the oversized anchor near the image border.
The default setting of the baseline model keeps all the anchors.
We remove all the anchors whose edges are outside the valid image.
The third is the image resolution.
The default resolution of the baseline model is $1333\times 800$, and we train a model with a small one, e.g., $640\times 640$.
The results are reported in Table \ref{tab:3factors}.
One can see that the three modifications result in a significant AP drop.
The full-conv padding can lower the ZP variance, but it is not helpful for detection accuracy.
In addition, it cannot obtain a better spatial equilibrium by removing the out-of-bounds anchors or setting different image resolutions,
because the supervision signal remains imbalanced between the zones.
It is challenging to find a solution that can alleviate spatial disequilibrium problem without causing performance degradation.

\section{Conclusions, Challenges and Outlook}\label{sec:conclusion}

In this paper, we present zone evaluation to reveal the existence and the discrete amplitude of spatial bias in modern object detectors.
We find that the spatial bias is less relevant to the object scale and the absolute positions of objects, but closely related to the gap of object data patterns between the zones.
Based on the thorough study of the origin of spatial bias, we finally introduce the spatial disequilibrium problem, aiming at a robust detection across the zones.
We also show a path to spatial equilibrium object detection, as a start to alleviate this problem.
Extensive experiments manifest the existence and the major source of spatial bias, which is prevalent in various modern detectors and datasets.

\textit{Significance.}
Spatial bias is a natural obstacle in object detection that the detectors usually show a performance drop in the border zone, which occupies a large proportion of the image area.
While the classic AP metric is still considered to be the primary measurement, it is difficult to reveal spatial bias and is challenging to comprehensively reflect the real performance of object detectors.
Maximizing the AP metric does not fully indicate a robust detection and performs well in all zones.
Zone evaluation supplements a series of zone metrics, compensates for the drawbacks of traditional evaluation, and captures more information about detection performance.
We hope this work could inspire the community to rethink the evaluation of object detectors and stimulate further explorations on spatial bias, and the solutions to the spatial disequilibrium problem.

There are several challenges left behind in this work:

\begin{table}[!t]
  \footnotesize
  \centering
  \setlength{\tabcolsep}{3pt}
  \caption{Evaluation of 3 potential factors for spatial bias. (1) We use full-conv \cite{kayhan2020translation} padding in the head network; (2) We remove the oversized anchor boxes that exceed the valid image border; (3) We set the image resolution as $640\times 640$. 5 zone precisions (ZP), the variance of ZP, and the traditional metric AP are reported. The results are reported on GFocal \cite{gfocal} on COCO val2017.
  }\vspace{-0pt}
  \begin{tabular}{cccccccc}
	Modification & AP  & Var. & ZP$^{0,1}$ & ZP$^{1,2}$ & ZP$^{2,3}$ & ZP$^{3,4}$ & ZP$^{4,5}$ \\ \hline
    Baseline & 40.1 & 16.9 & 31.1 & 37.5 & 39.4 & 38.5 & 43.8 \\
    \hline
    (1) & 38.6 & 12.4 & 30.4 & 36.0 & 37.5 & 37.8 & 41.2 \\
    (2) & 38.5 & 16.7 & 28.8 & 35.8 & 37.8 & 37.2 & 41.3 \\
    (3) & 36.9 & 18.8 & 26.7 & 33.6 & 36.2 & 34.9 & 39.9 \\
  \end{tabular}
  \label{tab:3factors}
  \vspace{-5pt}
\end{table}

\textit{Interpretability of spatial bias in various object detectors.}
This paper mainly reveals the existence and the discrete amplitude of spatial bias in object detectors, whereas the specific reason why different detectors perform quite differently is still frozen in the ice.
The neural network architecture designs, pre-training data, optimization, training strategies, and even hyper-parameters may play a role in the spatial bias.
Further exploration to answer the above question is of paramount importance.

\textit{The effect of other potential factors on the spatial bias.}
Currently, we pinpointed an evident correlation between imbalanced object distribution and zone performance.
There are some complicated yet implicit factors such as image blur, object occlusion, border effect, noise, etc., that may also contribute to spatial bias.
However, current detection datasets almost lack such annotations for the above factors, making it difficult to establish a quantitative analysis.

\textit{Zone evaluation for other vision tasks.}
Researchers have found some clues that the image generator may generate distorted content near the image border \cite{choi2021toward}.
Hence, the spatial bias may also exist in many vision tasks.
Our zone evaluation may have great potential to reveal spatial bias, whether for high-level or low-level vision tasks.

{
\small
\bibliographystyle{plain}
\bibliography{egbib}
}

\end{document}